**Towards improving Alzheimer's intervention: a machine learning approach for biomarker detection through combining MEG and MRI pipelines**

Alwani Liyana Ahmad[1,2,3], Jose Sanchez-Bornot[4, *], Roberto C. Sotero[5], Damien Coyle[6], Zamzuri Idris[2,3,7], Ibrahima Faye[1,8, *]

[1] Department of Fundamental and Applied Sciences, Faculty of Science and Information Technology, Universiti Teknologi PETRONAS, Perak, Malaysia.
[2] Department of Neurosciences, Hospital Universiti Sains Malaysia, Kelantan, Malaysia.
[3] Brain and Behaviour Cluster, School of Medical Sciences, Universiti Sains Malaysia, Kelantan, Malaysia
[4] Intelligent Systems Research Centre, School of Computing, Engineering and Intelligent Systems, Ulster University, Magee campus, Derry~Londonderry, BT48 7JL, UK.
[5] Department of Radiology and Hotchkiss Brain Institute, University of Calgary, Calgary, AB, Canada.
[6] The Bath Institute for the Augmented Human, University of Bath, Bath, BA2 7AY, UK.
[7] Department of Neurosciences, School of Medical Sciences, Universiti Sains Malaysia, Kelantan, Malaysia
[8] Centre for Intelligent Signal & Imaging Research (CISIR), Universiti Teknologi PETRONAS, Perak, Malaysia.

## Abstract

**Background:** Magneto/eletro-encephalography (MEG/EEG) are non-invasive neuroimaging techniques that offer excellent temporal and spatial resolution that can be used to study brain functioning. It has been proven effective in research on dementia and Alzheimer's disease (AD), where MEG/EEG analyses can identify changes in brain activity at various stages of AD progression, including preclinical and prodromal. Research has shown differences in the power spectrum properties among healthy controls (HC) and individuals with mild cognitive impairment (MCI) or AD. It has also been conjectured that MEG/EEG analyses have the capability to identify pathological changes before first clinical symptoms and may provide biomarkers for interventions designed to reverse or slow the progression of AD.

**Objective:** To conduct a comprehensive analysis by evaluating the performance of various classification techniques using features extracted by following different MEG methodologies. This study will focus on selecting MEG-based biomarkers/features that enhance performance in distinguishing between healthy control (HC) and mild cognitive impairment (MCI) participants from the BioFIND study. Furthermore, we will compare this with the effectiveness of using MRI-based anatomical features extracted from same participants, independently or in combination with the MEG features; thus, extending from previous study where similar analyses were performed using ADNI and OASIS-3 datasets.

**Methods:** MEG and MRI data are obtained from the BioFIND dataset, comprising 324 participants with 158 individuals diagnosed as MCI and 166 HC. Sensor and source-based MEG analyses are performed using custom MATLAB code based on SPM12 and OSL toolboxes, for co-registration, preprocessing including filtering, and automatic artifact rejection. Source-based analyses were applied through different source localization methods as implemented in SPM12 and DAiSS toolboxes. We conducted analogous machine learning analyses on sensor and source spaces, both separately for the MEG magnetometer (MAG) and gradiometer (GRAD) signals and with/without applying a data-correction or harmonization procedure over the extracted features. For the MEG-based features, this procedure followed from the BioFIND white paper's original correction approach, consisting of removing the effects of age, recording site, and other nuisance variables. Moreover, 100 Monte-Carlo replications of $K$=10-folds nested cross-validation partitions were performed to evaluate the robustness of classifiers performance and selected features through N-way ANOVA analyses. For each analysis, the dependent variable corresponds to the different performance statistics (Acc – Accuracy, Sens – Sensitivity, Spec –

Specificity, AUC – Area Under the Receiver Operator Curve) measured under different factors. As example for one of the analyses, the factors corresponded to: (i) sensor- or source-based analyses, where the latter were implemented independently for different source localization methods; (ii) considering the original or harmonized features; (iii) obtained either from the MAG or GRAD signals; and (iv) for different machine learning approaches, e.g., Gaussian Naïve Bayes (GNB), Kernel Support Vector Machine (KSVM) and Logistic Regression with L1 penalty (GLMNET).

**Results:** The best performance was achieved by combining altogether the MRI features with the features extracted from source-based MAG and GRAD signal analyses, which were obtained based on the Linearly Constrained Minimum Variance (LCMV) inverse solution, with an accuracy of 76.31% and AUC of 0.82 achieved for the GLMNET classifier for uncorrected MEG features and z-score-corrected MRI features. In contrast, the highest performance using exclusively only the features extracted from MEG's MAG or GRAD analyses, while combining with the MRI's, was obtained for the source-based GRAD signal analysis based on eLORETA inverse solution, with an accuracy of 75.46% and AUC of 0.81 achieved for uncorrected MEG and z-score-corrected MRI features from GLMNET. Furthermore, using only features from a single modality, classification analysis based on the LCMV inverse solution of MAG signals outperformed the rest, including MRI-based analyses. The former achieved an accuracy of 74.77% and AUC of 0.81 for the KSVM classifier, whereas for only z-score-corrected MRI features the best performance was achieved by the GLMNET classifier, with an accuracy of 72.74% and AUC of 0.79. Overall, the analyses revealed that features extracted from LCMV and eLORETA solution were consistently producing superior results (p-value $p < 10^{-7}$), while KSVM and GLMNET were the top classifiers ($p < 10^{-7}$). Based on the results, we strongly recommend using z-score data normalization for the MRI features ($p < 10^{-3}$) and no correction for the MEG features ($p < 10^{-3}$).

**Conclusion:** Despite the availability of multiple options, e.g., among different inverse solution and classification methods, we provided a comprehensive approach that combines MEG and MRI features. The use of Monte-Carlo analysis consisting of 100 replications of a 10-fold nested cross-validation procedure was critically important to realise the danger of overfitting and evaluate the "better" techniques. Even when some of the explored classifiers implement different hyperparameter evaluation strategies, we observed that overfitting can appear even in very tightly controlled scenarios, and therefore it is essential to validate the outcome with unseen data, as implemented in our study through nested cross-validation. The classifiers that have shown superior performance are KSVM and GLMNET. Finally, it is shown that exclusively using only MEG features can contribute to comparable or higher performance than using strictly only MRI features, with the LCMV and eLORETA inverse solutions allowing for better performance among compared approaches; however, as also demonstrated, combination of source-based uncorrected MEG and z-score-corrected MRI features is preferable.

## 1.0 Introduction

Alzheimer's disease (AD) is characterized by underlying physiological changes affecting the brain white and grey matter tissues which leads to the dysfunction of local and global neuronal networks, which in turn is reflected in the impairment of memory and cognitive functions (Sanchez-Rodriguez and Medina 2023). Neuroimaging techniques such as magneto/electro-encephalography (MEG/EEG) are increasingly used to study the associated functional changes (Sylvain Baillet 2017; Schoffelen and Gross 2009; Darvas et al. 2004; Bénar et al. 2021; Tagliazucchi and Laufs 2015), where alterations in oscillatory activity have been observed in individuals with AD. These alterations indicate that network disruptions could be detected early in the disease's trajectory, including during the mild cognitive impairments (MCI) or earlier stages (Yook 2022). For example, the use of power spectral density (PSD) analysis is commonly employed to identify abnormal neuronal activity. Several studies have shown that the relative power spectrum and spectral entropies in different frequency bands is altered in Alzheimer's patients with respect to control subjects (Aggarwal and Ray 2023; Verma, Lerner, and

Mizuiri 2023). The main findings point to amplified power in the theta and reduced power in the alpha band (Poza et al. 2007; Sanchez-Bornot et al. 2021).

Studying the brain resting state activity through MEG/EEG analyses have been widely utilized to investigate neuronal dynamics in healthy aging, MCI and AD (Mandal et al. 2018; Scheijbeler et al. 2022; Hughes et al. 2019; Yang et al. 2019; Sanchez-Bornot et al. 2021). Although it is crucial to ensure that subjects remain awake and alert, as drowsiness can significantly impact the brain activity (Verdoorn et al. 2011), employing an eyes-closed resting state is most advisable to mitigate eye movement and enhance the reproducibility of studies across multiple MEG/EEG recording sites. Research have shown that individuals with amnestic MCI (aMCI) exhibit a slowing of brain activity, characterized by a decrease in power for high frequency oscillations, including alpha and beta bands, and an increase in power in theta oscillations (Bruña et al. 2023; Wang, R., Wang, J., Yu 2015). These techniques have also been used to distinguish individuals in early AD stages from healthy controls (HC), with alpha relative power in the source space and spectra ration, particularly alpha/beta power ratio, being employed as biomarkers (López-sanz et al. 2019; Davenport 2023).

Importantly, MEG/EEG-based brain source imaging is a non-invasive technique which allows the estimation of neuronal current densities, and thus enabling to study functional effects of brain atrophy or less dramatic changes, as may be expected to occur in early cognitive decline. When neurons in about a 1.5 mm$^2$ patch of the brain are activated synchronously, mainly pyramidal cells, it produces steadily oriented ionic currents with associated electromagnetic fields that can be detected outside of the scalp (Koponen and Peterchev 2020). The equivalent current dipole (ECD) model simplifies the biophysical mechanisms of the generation of neuronal electromagnetic activity (Mosher, Leahy, and Lewis 1999), making it possible to numerically assess the contribution of these currents on recorded MEG/EEG signals by solving electromagnetic forward problems/equations (Hämäläinen, Huang, and Bowyer 2020). Inverting those equations, at least it makes practical to estimate the location and orientation of surrogate neuronal sources after the recording of MEG/EEG signals (Hämäläinen, Huang, and Bowyer 2020; Velmurugan, Sinha, and Satishchandra 2014). However, these solutions are not exempt of issues emanating from the ill-conditioning of brain inverse problems and volume conduction (Sanchez-Bornot et al. 2022), where the latter translates into the high correlation of measured activity, especially between nearby recording sites. Remarkably, both MEG/EEG have excellent temporal resolution (~1 ms), but it is considered that MEG allows superior spatial resolution for estimating cortical activity (Pantazis and Leahy 2006). Otherwise, EEG is cheaper and more accessible and portable. Consequently, MEG/EEG-based source neuroimaging is considered potentially an essential technique for close-up examination of normal and abnormal brain activity and advancing the development of brain diagnostic tools (Hämäläinen et al. 1993; Fergus et al. 2016; Dafflon et al. 2020; Lv et al. 2021; Tandel et al. 2019; Singh and Singh 2021).

Limitations of MEG/EEG inverse solution methods make challenging the analyses of normal/abnormal brain activity. This is more evident when studying brain functional connectivity, where the selection among the many available source localization approaches directly impact the connectivity assessment (S. Baillet, Mosher, and Leahy 2001; Grech et al. 2008; Mahjoory et al. 2017; Hincapié et al. 2017). Therefore, it is an open problem in neuroscience research how to choose the most appropriate inverse solution (Vallarino et al. 2023; Asadzadeh et al. 2020).  Similarly, there are many options or pipelines for the preprocessing of MEG/EEG signals and extracting the needed biomarkers/features for classification analysis (Kabir et al. 2023; Molla et al. 2020).

In our study, we will use five source localization approaches, as available in the SPM12 and DAiSS toolboxes, in addition to different preprocessing choices such as extracting biomarkers from MEG's magnetometer (MAG) and gradiometer (GRAD) signals, separately, as well as from sensor- and source-based analysis and from the unprocessed or data-corrected features. The use of multiple inverse solution methods could ameliorate the effects of the ill-conditioning of inverse problems and volume conduction on the quality of extracted brain activity features or biomarkers in the sense that some methods could be less sensitive to those issues and therefore will enable extracting better biomarkers. Using the BioFIND dataset (Delshad Vaghari, Ricardo Bruna, Laura E. Hughes, David Nesbitt, Roni Tibon, James B. Rowe, Fernando Maestu 2022), we will also evaluate different feature combinations among the MEG and magnetic resonance imaging (MRI) extracted features for the same participants and use multiple classification methods to improve the assessment of biomarkers. Critically, we will provide a classification pipeline implementing a Monte Carlo (MC) evaluation of nested $K$=10 folds cross-validation to avoid the overfitting problem in classification analysis and perform a methodically evaluation of the results. Our goal is to achieve robust outcomes and superior classification performance through this computationally demanding methodological strategy. Ultimately, this methodology combined with the use of N-way ANOVA analysis will contribute to extract critical information about possible better choices for preprocessing, source localization and classification approaches, as well as for potential more effective biomarkers.

## 2.0 Methodology

Our methodology (**Fig. 1**) involves the preprocessing and analysis of MEG and MRI data. In the case of MEG signals, two different pipelines are implemented for extracting features in sensor and source space, separately, MEG's MAG and GRAD signals. The features correspond to relative power spectrum in delta ($\delta$), theta ($\theta$), alpha ($\alpha$), beta ($\beta$), and low/high gamma ($\gamma$) frequency bands. Classification analysis for different machine learning approaches was conducted independently for MEG- and MRI-based features, and for different features combination, using a pipeline implementing 100 MC replications for $K$=10 fold nested cross-validation. The 100 performance outcomes of this pipeline for different stats (Acc – accuracy, Sens – sensitivity, Spec – specificity, and AUC – area under the receiver operator curve) are finally submitted to N-way ANOVA analyses to evaluate the best extracted features and pipeline choices while controlling for multiple comparisons.

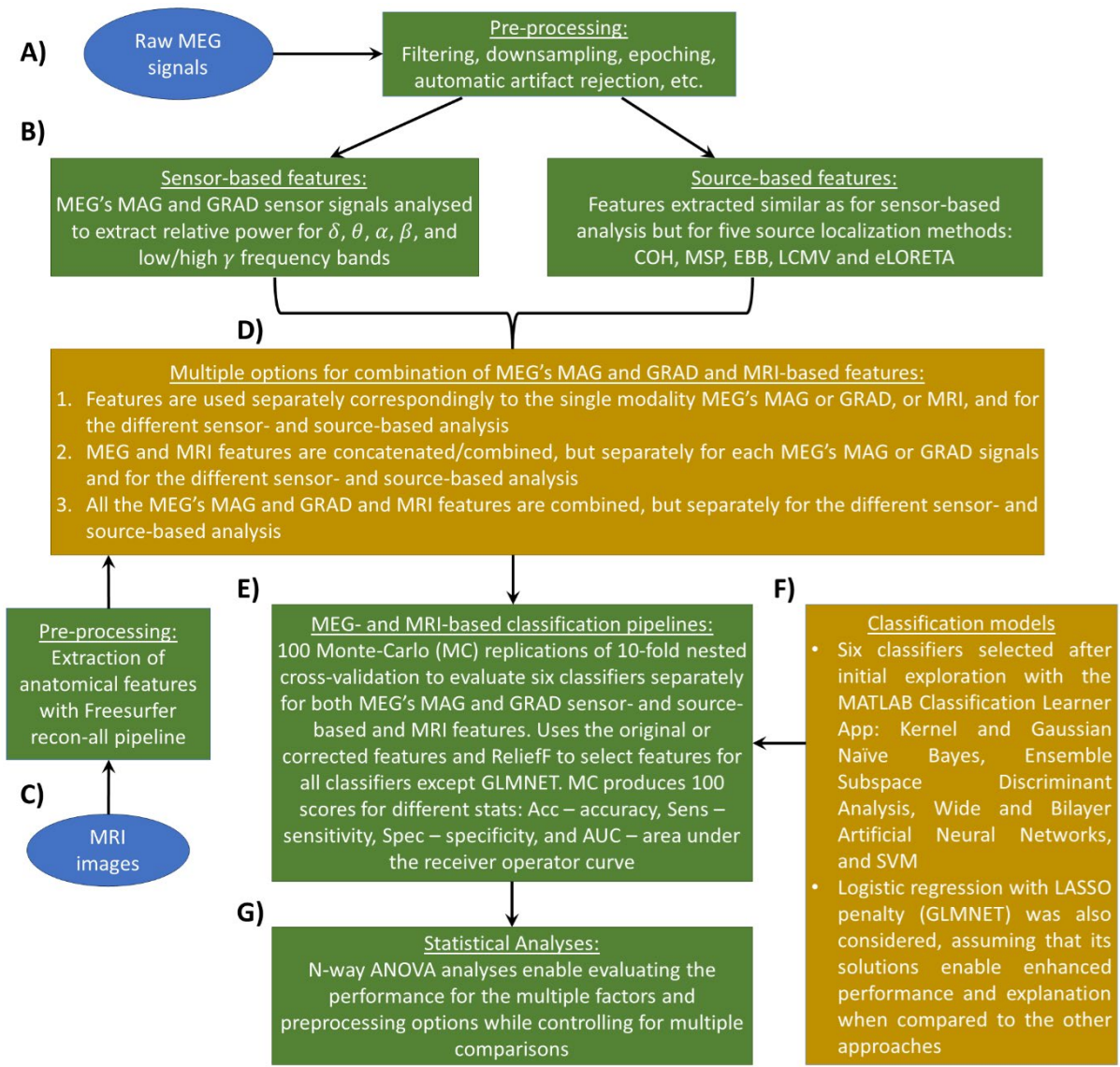

**Figure 1:** Workflow illustrating the proposed methodology. **(A)** MEG's magnetometer (MAG) and gradiometer (GRAD) are preprocessed using standard SPM12 and OSL tools pipelines. **(B)** Sensor- and source-based features are extracted corresponding to relative power calculations for $\delta$, $\theta$, $\alpha$, $\beta$, and low/high $\gamma$ frequency bands. **(C)** MRI data is pre-processed using Freesurfer recon-all pipeline. **(D)** Multiple options are considered for combining MEG's MAG and GRAD and MRI features. **(E)** A customized Matlab pipeline implements Monte-Carlo nested cross-validation for robust evaluation of the tested classification methods **(F)** and evaluated features. **(G)** N-way ANOVA analyses are implemented to evaluate the best extracted features and tested approaches. The inputs for this methodology are represented with blue ellipses. The green rectangles represent the different implemented pipelines. The orange rectangle represents some of the multiple evaluated options, including features combination and different tested classifiers.

## 2.1 Participants data

The BioFIND dataset provides multisite MEG and MRI data for 324 participants (166/158 HC/MCI) as summarised in **Table 1**. The data were collected in the UK Cambridge's Cognitive & Brain Sciences Unit (CBU) and the Spain Madrid's Centre for Biomedical Technology (CTB). In both cases, MEG recordings were continuously collected using Elekta Neuromag Vector view 306 MEG systems sampled at a rate of 1 kHz and were acquired during resting state eyes-closed condition. Despite the same recording modality and paradigm, the signals are heterogeneous as the study comprises at least two different populations and different controlled and uncontrolled scenarios. For example, the CBU magnetic shielding room (MSR) was made by Imedco and uses single layer mu-metal plates, while the CTB MSR was made by Vacuumschmelze and has two layers. Moreover, for the CBU, the average MSR noise level during tuning was 2.3 fT/sqrt (Hz); for the CTB, it was 2.8 fT/sqrt (Hz) until 2016, and 2.6 fT/sqrt (Hz) after 2016. Therefore, the BioFIND dataset is an excellent resource to evaluate the combination of MEG data across multiple sites to study AD as a proof for future studies.

Together with the MEG data collected for each participant, for most of them also T1-weighted MRI data was acquired at both sites. CBU's MRIs were obtained using either a Siemens 3T TIM TRIO or Prisma scanner, employing an MPRAGE pulse sequence; whereas the CTB's MRIs were acquired using a General Electric 1.5 Tesla MRI system, with a high-resolution antenna equipped with a homogenization PURE filter.

**Table 1:** Summary of data characteristics. **Mean (SD) shown for Age, Education and MMSE

| Data Characteristics | HC | MCI |
|---|---|---|
| Site (UK's CBU/Spain's CTB) | 91/75 | 68/90 |
| Gender (M/F) | 82/84 | 80/78 |
| Age (years) ** | 71.3 (7.0) | 72.9 (6.7) |
| Education (years) ** | 14.5 (4.4) | 10.8 (5.3) |
| MMSE ** | 28.8 (1.2) | 26.1 (2.8) |

## 2.2 MEG features extraction

The statistical analyses and data access were performed on the Dementia Platform UK (DPUK)'s, which is accessible through https://portal.dpuk.ukserp.ac.uk/ as stipulated in the signed Memorandum of Understanding (MOU) documents. DPUK provided virtual desktops which allowed to implement and run the analyses on their virtual premises. The available MEG data was acquired using the Elekta Neuromag system which includes two different sensor types: 102 MAG and 204 GRAD channels. MAG sensors are more sensitive to the component of the magnetic field that is perpendicular to the sensors array, whereas GRAD sensors electronically calculate the spatial derivative of the magnetic fields along two orthogonal directions in the tangential plane to the sensor array (Hämäläinen et al. 1993). GRAD's capability to calculate spatial derivatives enables diminished sensitivity towards distant sources, such as environmental noise. Thus, enhancing the recorded signal-to-noise ratio for signals generated from nearby cerebral cortex sources. In contrast, MAG sensors are also sensitive to deeper brain sources and other extra-brain signals, including ambient noise. As combining MAG and GRAD data can be challenging due to different signals and noise scales (MAGs; fT while GRADs; fT/cm) (Westner et al. 2022; Jaiswal et al. 2020a; Garcés et al. 2017b), we performed separated statistical analyses for these signals at sensor as well as at source level.

The available MEG data consist of 5 minutes resting-state recordings with eyes closed. The frequency bands used in our analyses are $\delta$ (2-4 Hz), $\theta$ (4-8 Hz), $\alpha$ (8-12 Hz), $\beta$ (12-35 Hz), low $\gamma$ (30-48 Hz) and high $\gamma$ (52-86 Hz). Data pre-processing, including filtering and downsampling to 250 Hz, was performed using the Statistical Parametric Mapping (SPM12) toolbox (William D. Penny, Karl J. Friston, John T. Ashburner, Stefan J. Kiebel 2011), while automatic artifact rejection was performed using the OSL toolbox (https://ohba-analysis.github.io/osl-docs/). Particularly, we applied a low-pass 9th-order Butterworth filter with cutoff frequency of 95 Hz to remove the strong artefactual effects of head position indicators (HPI) coils, energized around 150 Hz and 300 Hz in the CTB and CBU recordings, respectively. Successively, we applied 4th order Butterworth high-pass and notch filters, with cutoff frequency of 1 Hz and stop band of 49–51 Hz, respectively. Lastly, we extracted 1 sec length non-overlapping epochs and automatically removed epochs with artifacts, such as generated by eyes and muscular movements, using available OSL tools, resulting in around 115 ± 7.4 epochs per participant data.

While for sensor-based analyses, the sensor signals were directly used after the above pre-processing steps, for source-based analyses these were projected to brain space using five different inverse solutions as available in SPM12 (COH – Bayesian LORETA, MSP – Multiple Sparse Priors, and EBB – Empirical Bayesian Beamformer) and its DAiSS plugin (Jaiswal et al. 2020b) (eLORETA – exact Low Resolution Tomography Analysis, and LCMV beamformer – Linearly Constrained Minimum Variance). For projection on the brain cortical surface, we used 500 dipoles obtained from calculating a *K*-medoids partition for the 8192 vertices of the medium-size cortical surface provided in SPM12, after co-registration with the participant's individual brain anatomy. Remarkably, the use of different source localization methods and comparing the classification performance obtained from their independently derived features could serve as new evidence to support the more appropriate inverse solution approaches.

As aforementioned, sensor- and source-based analyses were performed separately for the MEG's MAG and GRAD signals. Thereby, for the MAG analysis in the sensor space, the number of features is 102 sensors x 6 frequency bands = 612 features. Similarly for the GRAD analysis, the number of features is 204 x 6 = 1224 features. For both source-based analyses based on MAG and GRAD signals, the number of features is 500 x 6 = 3000 features. Lastly, these outcomes from either sensor- or source-based analyses, from MAG or GRAD signals, were further processed using a data-correction procedure to remove nuisance variable effects, such as caused by the uncontrolled differences in recording site (CBU code=1, CTB code=2) and tracked head movements (Vaghari, Kabir, and Henson 2022), or left unprocessed to be used as biomarkers for the post-hoc classification analyses.

## 2.3 MRI features extraction

The MRI images were acquired in their NIFTI format and then processed using Freesurfer software package (surfer.nmr.mgh.harvard.edu), as available in the DPUK virtual environments (version: freesurfer-linux-ubuntu18_x86_64-dev-20220419-7af6446) with the standard cross-sectional pipeline *recon-all*. The primary processing steps conducted by Freesurfer included removing of non-brain tissue (stripping), by utilizing a hybrid watershed/surface deformation procedure to eliminate non-brain elements from the images (Ségonne et al. 2004), automated Talairach transformation, through an automated process to align the MRI scans with the Talairach atlas ensuring accurate spatial positioning and registration (Lancaster et al. 2000), and performing segmentation of subcortical white matter and deep grey matter volumetric structures (Bruce Fischl, David H. Salat, André J.W. van der Kouwe, Nikos Makris, Florent Ségonne, Brian T. Quinn 2004). Additionally, a measure of total intracranial volume (ICV) is estimated as part of the processing pipeline. This ICV calculation provides an assessment of the overall volume within the skull, serving as a valuable metric as a control reference for analyzing brain structural changes across development, healthy aging, and disease (Popuri et al. 2020; Ma D, Popuri K, Bhalla M, Sangha O, Lu D, Cao J, Jacova C, Wang L 2019; R. A. I. Bethlehem, J. Seidlitz, S. R. White, J. W. Vogel, K. M. Anderson, C. Adamson, S. Adler, G. S. Alexopoulos, E. Anagnostou, A. Areces-Gonzalez, D. E. Astle, B. Auyeung, M. Ayub, J. Bae, G. Ball, S. Baron-Cohen, R. Beare, S. A. Bedford, V. Benegal, 2022).

For harmonization of measures calculated with the Freesurfer pipeline, we adopted a multivariate polynomial regression approach, complementarily to previous studies, removing the nuisance effect of age, gender and total intracranial volume (TIV) over measures (Popuri et al. 2020; Ma D, Popuri K, Bhalla M, Sangha O, Lu D, Cao J, Jacova C, Wang L 2019; Ledig et al. 2018; Koikkalainen et al. 2012). Two different data correction procedures were implemented: 1) simply selecting the residuals after polynomial interpolation as corrected features, and 2) additionally calculating the variance at each interpolation point to transform the original measures into z-score-corrected features.

## 2.4 MEG- and MRI-based classification pipelines

To implement the classification pipeline, we first performed an exploratory analysis using the MATLAB R2022b Classification Learner (MCL) app using pre-processed MEG data. This step is designed to simplify the process of building and fine-tuning classification models and select a reduced number of methods for further analyses. From the MCL app, we explored all the available algorithms, including decision trees, discriminant analysis, logistic regression, kernel and gaussian naïve Bayes, support vector machine (SVM), nearest neighbours, kernel approximation, ensemble methods, and neural networks. We used the ReliefF method for feature selection, selecting only the features with positive scores since near zero or nonpositive scores indicate less important predictors (Kira et al. 1992), while the data was partitioned into train and test subsets randomly (80%/20% partition). This process was repeated 10 times to ensure reliability in the results and a fair selection of model candidates for posterior analyses. For consistency, when comparing with the results among the analyses using the MEG and MRI features, we finally selected the same classification models among the best performers but also considering the best classifiers observed from our previous study (Ahmad et al. 2024). In summary, we selected six models among the best performers: Kernel and Gaussian Naïve Bayes, Wide and Bilayer Artificial Neural Networks, Ensemble Subspace Discriminant Analysis, and Kernel Support Vector Machine (KSVM). Additionally, we used in our pipeline the GLMNET tool, as provided in its more advanced implementation in the R's package (J. Friedman, Hastie, and Tibshirani 2010; A. J. Friedman et al. 2010). GLMNET is used because of its flexibility to perform both feature selection and classification simultaneously, apart from excellent results reported in the literature (Kang et al. 2019).

Moreover, we implemented a classification pipeline based on multiple options among 1) classification methods; 2) feature signal modality: MEG's MAG or GRAD, or MRI, with the added characteristic that features extracted from MEG signals can be extracted from sensor- or source-based analysis; 3) with the consideration of using the original or harmonized features. Remarkably, we implemented nested $K$=10-fold cross-validation to avoid overfitting and repeated this procedure within a Monte-Carlo (MC) analysis designed to produce robust assessment of quality measurements for each classification outcomes (Acc − Accuracy, Sens − Sensitivity, Spec − Specificity, AUC − Area Under the Receiver Operator Curve). Because the nested cross-validation procedure creates a random partition of the samples, then each run can produce different outcomes. Therefore, this procedure was repeated using 100 MC replications to obtain a population of outcomes that will be later utilized by ANOVA analyses. Except KSVM and GLMNET, the other classification methods lack hyperparameters, and thereby are more prone to show overfitting issues. In the case of KSVM, the hyperparameters correspond to the type of kernel function, etc., which selection was evaluated using MATLAB Bayesian optimization tools. In the case of GLMNET, the single hyperparameter, which controls feature selection (L1 norm penalty), was selected within the cross-validation step, after following the solutions path calculated by GLMNET. As shown in the pseudo-code in **Table 2**, the implemented "nested" strategy creates a partition of 10% holdout and 90% test samples for the external $K = 10$ folds cycle, and the test samples are further partitioned within an internal $K − 1$ folds cycle into validation and train samples (10% and 80% with respect to the total number of samples, respectively) which are used to calculate the classification model parameters and assess hyperparameter values within a standard cross-validation procedure (Parvandeh et al. 2020).

This way, the statistical samples calculated for the 10% holdout data are averaged while rotating the holdout data for each of the $K = 10$ random partitions. By multiplying this process by the number of MC iterations, we implement an unbiased method to truly evaluate the classification models and selected features. Finally, the outcomes from the classification pipeline (100 MC measures for Acc, Sen, Spec, and AUC, for each classifier and methodology options) were submitted to N-way ANOVA analyses for an overall evaluation and report of the results, including assessments about possibly better inverse solution and classification methods, as well as the evaluation of MEG and MRI feature combination strategies, among other interesting investigations, while controlling for multiple comparisons.

**Table 2:** Pseudo-code for the classification analysis pipeline. The inputs are the classification model (Model), data structure including both the predictors matrix and response vector (Data), a flag to apply *a priori* feature selection or not (ffsel), and a flag to apply data harmonization (fharm). If harmonization is applied (fharm is True) then it must be input also the type of harmonization transform ("residuals" or "z-score") and the nuisance variables or covariates, else a z-score transform is applied for each predictor. The difference between this last transform and the harmonization's z-score transform is that the latter is applied using polynomial interpolation with respect to the provided covariates (see **Materials and Methods**).

| |
|---|
| **Inputs:** Model, Data, ffsel, fharm, type, covariates |
| **Output:** crossval_stats, holdout_stats |
| 01:    crossval_stats = [] |
| 02:    holdout_stats = [] |
| 03:    $K = 10$ |
| 04:    folds $\leftarrow$ partition(nrows(data), $K$) |
| 05:    **for** $(k = 1, \dots, K)$ |
| 06:        [holdout_Data, bus_Data] $\leftarrow$ split(Data, folds($k$)) |
| 07:        **if** (fharm is True) |
| 08:            [bus_Data, coefficients] $\leftarrow$ harmonization(bus_Data, type, covariates) |
| 09:            holdout_Data $\leftarrow$ apply_harmonization(holdout_Data, type, covariates, coefficients) |
| 10:        **else** |
| 11:            [bus_Data, coefficients] $\leftarrow$ zscore(bus_Data) |
| 12:            holdout_Data $\leftarrow$ apply_zscore(bus_Data, coefficients) |
| 13:        **end** |
| 14:        **if** (ffsel is True) |
| 15:            rank $\leftarrow$ relieff(bus_Data, 10) |
| 16:            indcols $\leftarrow$ find(rank > 0) |
| 17:            bus_Data $\leftarrow$ select_features(bus_Data, indcols) |
| 18:            holdout_Data $\leftarrow$ select_features(holdout_Data, indcols) |
| 19:        **end** |
| 20:        stats = [] |
| 21:        **for** $(l = 1, \dots, K; l \neq k)$ |
| 22:            train_Data $\leftarrow$ select_samples(bus_Data, folds($\{1, \dots, K\}\setminus\{k, l\}$)) |
| 23:            test_Data $\leftarrow$ select_samples(bus_Data, folds($l$)) |
| 24:            coefficients $\leftarrow$ train(Model, train_Data) |
| 25:            stats $\leftarrow$ [stats, eval(Model, coefficients, test_Data)] |
| 26:        **end** |
| 27:        stats $\leftarrow$ average(stats) |
| 28:        coefficients $\leftarrow$ select_best(Model, bus_Data, stats) |
| 29:        crossval_stats $\leftarrow$ [crossval_stats, eval(Model, coefficients, bus_Data)] |
| 30:        holdout_stats $\leftarrow$ [holdout_stats, eval(Model, coefficients, holdout_Data)] |
| 31:    **end** |
| 32:    crossval_stats $\leftarrow$ average(crossval_stats) |
| 33:    holdout_stats $\leftarrow$ average(holdout_stats) |

## 2.5 Feature selection strategies

Basically, our study employed two different feature selection strategies. Except for GLMNET, all the other classifiers used ReliefF for a preliminary selection of features. Thus, helping to reduce the dimensionality of the feature space before fitting the classification model. However, this strategy can be criticized since features may be selected in disconnection with their predictive power in the classification task. However, as illustrated next, the ReliefF provides a very insightful mechanism to select features that has some protection from this issue. In contrast, GLMNET performs feature selection and classification simultaneously, which can be advantageous as features performing better in the classification are retained.

### 2.5.1 Selecting features before evaluating the classification model: ReliefF

Set $\mathbf{X} \in \mathbb{R}^{M \times N}$ as the features matrix, containing $M$ observations or samples, and $N$ predictors or features. For the binary classification task, HC vs. MCI in our case, ReliefF's rankings for each feature $i = 1, \dots, N$ are calculated as follows:

1)  $R_i = R_i - \frac{1}{LJ} \sum_{l \in \mathcal{R}(M,L)} [\sum_{j \in \mathcal{H}(l)} g(i, \pmb{x}_l, \pmb{x}_j) + \sum_{j \in \mathcal{M}(l)} g(i, \pmb{x}_l, \pmb{x}_j)],$

where $\mathcal{R}(M, L)$ is the subset of $L$ random row indices of $\mathbf{X}$, selected by sampling with replacement, $\pmb{x}_l \in \mathbb{R}^N$ for $l \in \mathcal{R}(M, L)$ represents a particular random row of $\mathbf{X}$, $\mathcal{H}(l)$ and $\mathcal{M}(l)$ represent the $\mathcal{H}its$ and $\mathcal{M}isses$ subsets of row indices of $\mathbf{X}$ (their cardinality is $J = 10$ as used for running the ReliefF algorithm), correspondingly to the $J$ nearest neighbours of $\pmb{x}_l$ in the sample space, i.e., containing the indices of $J$ neighbour samples for the same (hits) and different (miss) class as $\pmb{x}_l$, and

$$g(i, \pmb{x}_l, \pmb{x}_k) = \begin{cases} \frac{|x_{li} - x_{ki}|}{\max\limits_{j \in \{1, \dots, M\}} x_{ji} - \min\limits_{j \in \{1, \dots, M\}} x_{ji}}, & \text{if the } i - \text{th feature is continuous} \\ 0, & \text{discrete feature and } class(\pmb{x}_l) = class(\pmb{x}_k) \\ 1, & \text{discrete feature and } class(\pmb{x}_l) \neq class(\pmb{x}_k) \end{cases}.$$

As illustration, **Fig. 2** shows the ReliefF's rankings calculated for the MRI features (upper-half bar plot), which were extracted using the Freesurfer pipeline, as shown from top to bottom in decreasing order of relevance. Therein, two of the most relevant features were selected to plot the observations as projected in the planar subspace of these features (bottom-left corner scatter plot), and the same operation was performed but for two of the more irrelevant features (bottom-right corner scatter plot). As it may be noticed in these representations, features with the highest ranking provide better separation between the two classes (HC vs. MCI). For the two selected more irrelevant features, the rankings are around zero. As it can be seen in the projection representation for more irrelevant features, there is a great overlap between HC and MCI observations. Therefore, it is difficult to find a clear separation boundary in these features subspace to improve discrimination. Moreover, for features with negative ReliefF rankings, interpretation is more obscure as it can be notice in **Eq. (1)** that the first term inside the square brackets is larger than the second one. This means that for the neighbourhood of $\mathcal{H}its$ and $\mathcal{M}isses$, on "average" a random sample is more closely surrounded by samples of the other class than from its same class when considering their measures for this feature, i.e., more foes than mates in the neighbourhood. In summary, often studies select only the features with positive ReliefF's rankings, as done in our study.

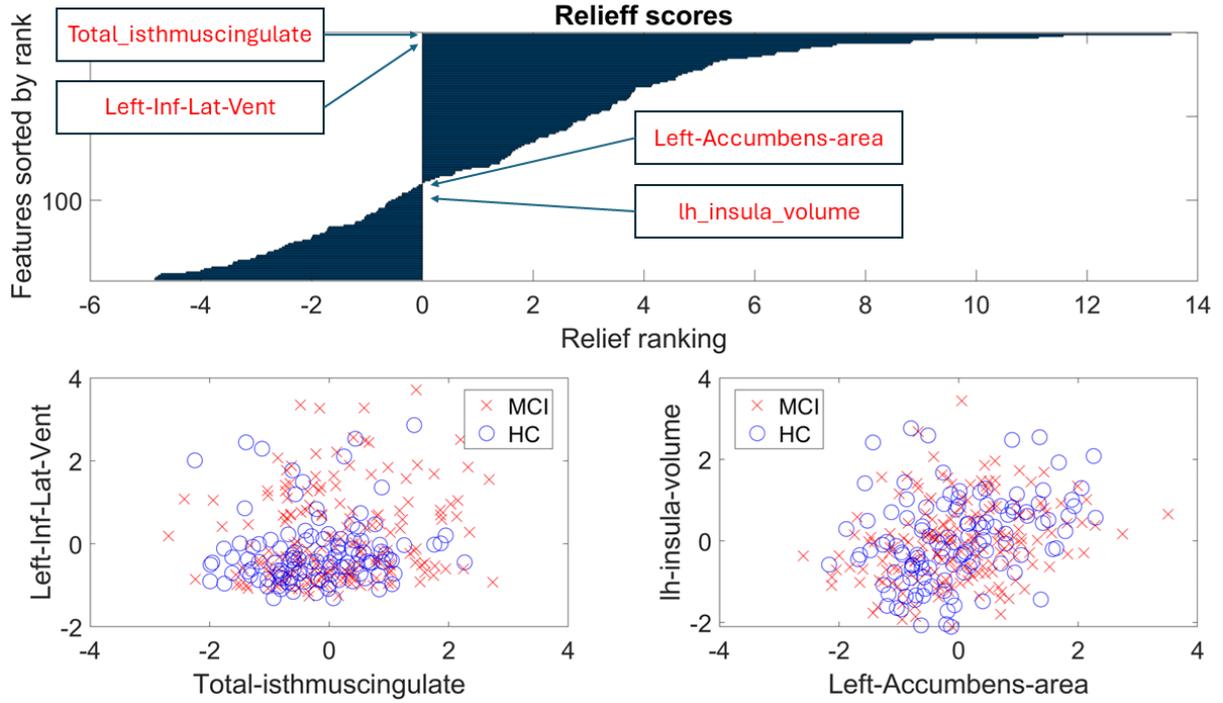

**Figure 2:** Feature ranking using ReliefF. Top half bar plot: rankings are sorted in descending order. Highlighted in enclosed rectangles are show two of the most relevant or other two less relevant features. Bottom half scatter plots: HC and MCI observation are plotted as projected in the subspace of these features separately for the more relevant (left) and less relevant features (right).

### 2.5.2 Interleaving classification model selection with feature selection: GLMNET

The logistic regression with LASSO regularization (called here as GLMNET) allows to select features while fitting the classification model. The general optimization problem solved by GLMNET can be stated as follows:

$$2) \qquad \hat{\beta} = \underset{\beta_0, \beta}{\operatorname{argmin}} \frac{1}{N} \sum_{i=1}^{N} w_i l(y_i, \beta_0 + \beta^T \mathbf{x}_i) + \lambda \left( (1 - \alpha) \frac{\|\beta\|_2^2}{2} + \alpha \|\beta\|_1 \right); \text{ for } \lambda > 0, \alpha \in [0; 1]$$

Here $l(\cdot)$ is the negative log-likelihood function, e.g., $l(\beta_0, \beta; y_i, z_i) = \frac{1}{2}(y_i - \beta_0 - \beta^T \mathbf{x}_i)^2$ in the Gaussian case and $l(\beta_0, \beta; y_i, z_i) = y_i(\beta_0 + \beta^T \mathbf{x}_i) - log\left(1 + e^{\beta_0 + \beta^T \mathbf{x}_i}\right)$ for our case in the logistic function. For using GLMNET, we selected $w_i = 1$, for $i = 1, \dots, N$, and $\alpha = 1$ in **Eq. (2)**, as we are strictly interested in the feature selection case; although, the $\alpha < 1$ case could be interesting in situations with many features or high correlation among them, as shown by elastic net regression (Tay, Narasimhan, and Hastie 2023). As implemented in the R's package (Hastie, Qian, and Tay 2023; A. J. Friedman et al. 2010), GLMNET can provide the "continuous" path of solutions corresponding to the hyperparameter $\lambda$, which is evaluated in the interval $(0; \lambda_{max})$.

### 3.0 Results

### 3.1 Preliminary model selection with MATLAB classification learner app

During the model selection process using the MCL app, the Ensemble Subspace Discriminant model consistently exhibited superior classification performance, appearing 24 times as the model with best classification performance, as shown in **Table 3**. It was closely followed by the Wide Neural Network model, which appeared 17 times. The KSVM and Bilayered Neural Network models also appeared frequently, with 12 and 11 occurrences, respectively. In addition, we incorporated two other classification approaches, Kernel and Gaussian Naïve Bayes models, which showed superior performance in our previous study with an Alzheimer Disease Neuroinformatic Initiative (ADNI)'s MRI dataset (Ahmad et al. 2024). Within the MCL app, the ReliefF method was used for feature selection, as discussed in the **Materials and Methods**. For the next analyses, these six classification models were joined by the logistic regression based on LASSO penalty (GLMNET) classifier, as an alternative method which performs simultaneously feature selection and classification. In contrast to the MCL app preliminary analysis, in the following we used the classification pipeline which provides a more robust statistical evaluation based on a Monte-Carlo (MC) replication analysis of nested *K*-fold cross-validation.

**Table 3:** Top 10 classification models with superior performance, as evaluated using the ReliefF method for feature selection and *K*=10 classical cross-validation approach, within the MATLAB Classification Learning (MCL) app.

| Models | Total |
|---|---|
| Ensemble: Subspace Discriminant | 24 |
| Neural Network: Wide Neural Network | 17 |
| Kernel SVM | 12 |
| Neural Network:  Bilayered Neural Network | 11 |
| SVM: Linear SVM | 9 |
| Neural Network:  Medium Neural Network | 8 |
| Neural Network: Narrow Neural Network | 4 |
| KNN: Fine KNN | 4 |
| Linear Discriminant | 4 |
| Ensemble: Subspace KNN | 3 |

### 3.2 Classification analysis exclusively based on MEG features

Within the classification pipeline analysis, which implements a MC analysis (100 replicas in our case) of *K*=10 nested cross-validation, we utilized seven models as aforementioned. In the case of KSVM, this pipeline performs Bayesian optimization instead of grid evaluation, which was used for GLMNET by automatically following the path of continuous solutions corresponding to values of $\lambda$ in the interval $(0; \lambda_{max})$. Distinctively from KSVM and GLMNET, the other five classifiers did not have any hyperparameter to fit. This is advantageous from the point of view of faster computations, but their results are prone to be overfitted because of the lack of hyperparameters controlling this issue. Despite of this limitation, some of these methods showed good performance. However, the best performances were observed for KSVM and GLMNET which are shown exclusively on the rest of the paper (see **Supplementary Materials** for results including for the whole set of tested classifiers).

For the analysis exclusively based on MEG features, the best accuracy was observed using the KSVM model for the uncorrected MEG's MAG features derived from the LCMV inverse solution, achieving an accuracy of 74.77% and AUC of 0.81 as shown in **Fig. 3**. Notably, the comparison of sensitivity performance between MAG and GRAD features using LCMV analysis and the KSVM model also showed that MAG features achieved the highest sensitivity at 71.39%, compared to 68.11% for GRAD features. This is an interesting finding considering that analyses involving features derived from GRAD signals typically enabled a higher performance. The overall analysis also demonstrates that the performance of GRAD signals is significantly better than that of MAG signals (Gross et al. 2013; Dash et al. 2021), with $F_{(1,16632)} = 1838.1$, $p < 10^{-7}$ (see highlighted in yellow in **Supplementary Material Table 1**). In contrast, a similar for MEG's GRAD derived features achieved an accuracy of 73.24% with AUC of 0.81. Moreover, we observed lower significant results when using the features derived from sensor space analysis. In this case, the highest accuracy was obtained from GRAD signals, also for the analysis based on KSVM classifier for uncorrected data, with an accuracy of 69.28% and AUC of 0.76. In contrast to

KSVM results, using GLMNET the highest performance yielded an accuracy of 72.94% and an AUC of 0.78, achieved by using uncorrected features obtained from LCMV-based source analysis of GRAD signals. Here, we observed a significant difference between the top performances for KSVM and GLMNET, with $p < 10^{-7}$ and 95% $CI = [0.0106; 0.0134]$.

As further evaluation, we conducted an N-way ANOVA analysis to examine the influence of the multiple factors on classification performance. These factors included the use of different classifiers, the impact of data correction, the type of MEG sensors, and the choice of source localization methods. To control for multiple comparisons, we selected the Bonferroni correction technique, as well as performed pairwise comparisons using the post hoc Tukey HSD test, as provided in the SPSS software. Based on the accuracy performance, uncorrected features demonstrated superior performance than using the correction procedure ($F_{(1,16632)} = 1695.7$ with $p < 10^{-7}$). Furthermore, during the assessment of possible superior source localization methods, the best performance was achieved with the LCMV solution, in contrast to the other approaches including sensor analysis ($p < 10^{-7}$ with 95% $C.I = [0.0409, 0.0434]$). Lastly, for the classifier comparison, KSVM was identified as superior ($p < 10^{-7}$ with 95% $C.I = [0.0106, 0.0134]$). In contrast, the Gaussian Naïve Bayes classifier displayed the least significant results, as well as MSP among inverse solutions, as detailed in **Supplementary Material Fig. 1** and **Table 1**.

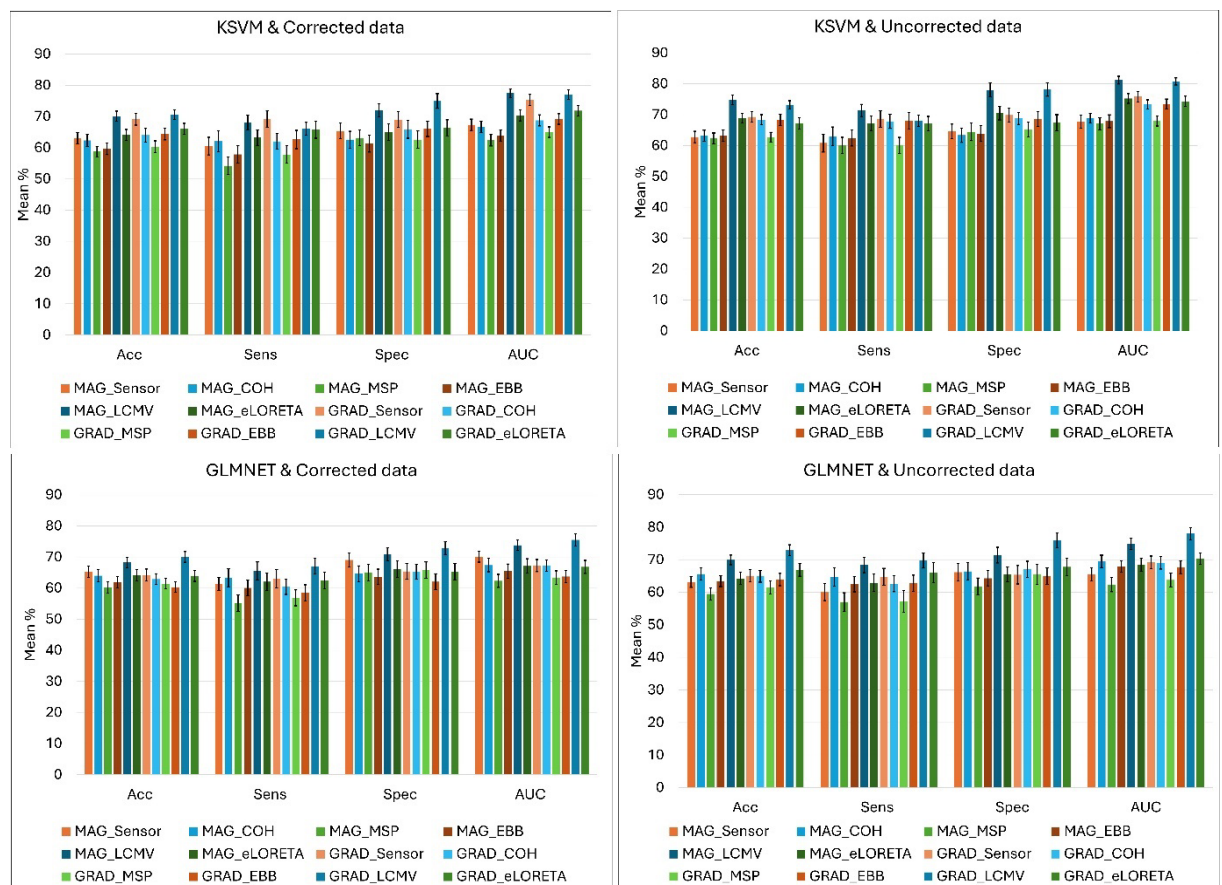

**Figure 3:** Comparison of classification performance using KSVM and GLMNET classifiers based only on MEG's MAG and GRAD derived features: (1) for corrected (left) vs. uncorrected (right) processing; and (2) for sensor and source level analyses, where the latter were implemented based on different inverse solutions: COH, MSP, EBB, LCMV and eLORETA.

### 3.3 Classification analysis exclusively based on MRI features

For MRI-only features based performance, the GLMNET model, utilizing z-score harmonization (Ahmad et al. 2024) (see **Materials and Methods**), achieved the highest results, with an accuracy of 72.74% and AUC of 0.79, as shown in **Fig. 4**. Here, the KSVM model also showed very good results for z-score harmonization with an accuracy of 71.75% and AUC of 0.79. In this case, in contrast to the analysis with MEG features, GLMNET significantly outperformed KSVM ($p < 10^{-3}$ with 95% $C.I = [0.0592; 0.0674]$), as shown in **Supplementary Material Table 2**. Moreover, in terms of data harmonization methods, the z-score significantly outperformed both uncorrected and residuals approaches ($p < 10^{-3}$).

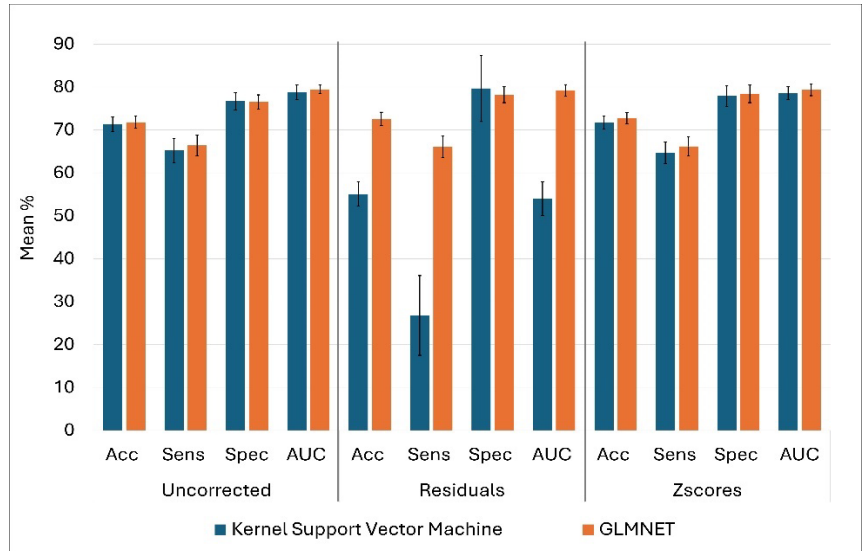

**Figure 4:** Classification performance exclusively based on MRI features for three different harmonization methods for only KSVM (blue bars) and GLMNET (orange) classifiers.

### 3.4 Combination of MEG and MRI features

In addition to above analyses, we employed two distinct approaches for combining MEG- and MRI-derived features. However, to reduce the number of analyses, we used only the uncorrected data from MEG features and z-score-corrected data from MRI features. In the first feature combination analysis, we combined MAG and GRAD features, separately, with MRI features. While, in the second approach we combined all together MAG and GRAD and MRI features. Notice that these combinations were done independently for sensor- and source-based analysis and for each of the evaluated inverse solutions.

For the first analysis, the highest performance was achieved using the GLMNET model for features obtained with eLORETA inverse solution of GRAD signals, achieving an accuracy of 75.46% and AUC of 0.81, as shown in **Fig. 5**. The second highest performance was achieved also for GLMNET using the eLORETA solution calculated from MAG signals, which produced a slightly lower accuracy of 75.35% and AUC of 0.81. However, over all the comparison N-way ANOVA analysis revealed that the LCMV outperformed eLORETA and the other source localization methods (e.g., for the LCMV vs. eLORETA pairwise contrast for accuracy performance: $p < 10^{-7}$ with 95% $C.I = [0.0124; 0.0189]$). Surprisingly, when assessing the overall performance between MAG and GRAD sensors, MAG sensors exhibited the highest accuracy performance ($F_{(1,8387)} = 109.9$ with $p < 10^{-3}$).

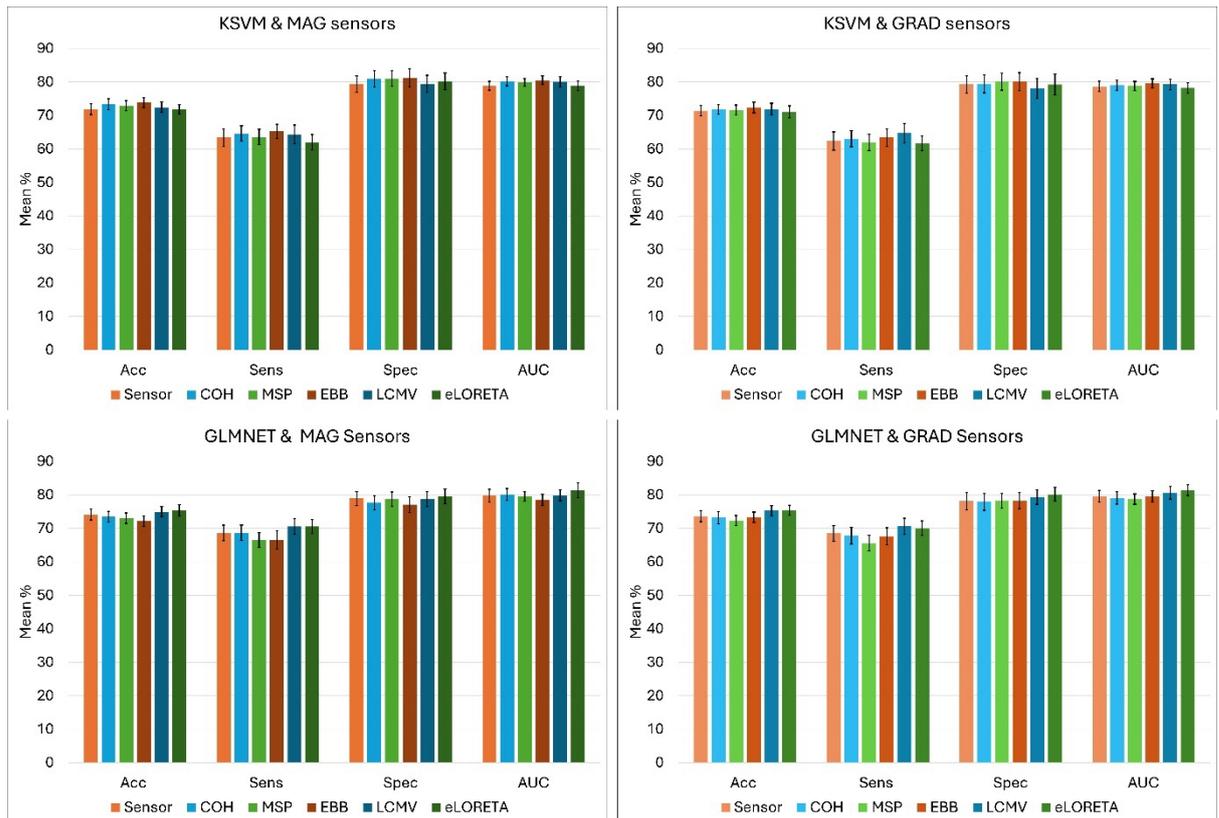

**Figure 5:** Comparative classification performance using KSVM (**top**) and GLMNET (**bottom**) models between outcomes for combining MEG's MAG with MRI features (**left**) and MEG's GRAD with MRI features (**right**), implemented separately for MEG sensor- and source-based analysis.

In the second approach, we combined both MAG and GRAD with MRI features. This strategy produced the highest performance using the GLMNET model for features derived from the LCMV inverse solution, achieving an accuracy of 76.31% and AUC of 0.82, as illustrated in **Fig. 6**. The second best result was also achieved by GLMNET but using eLORETA, with a lower accuracy of 75.45% and AUC of 0.81. Further comparative analysis of the overall performance based on accuracy showed that LCMV consistently outperformed eLORETA ($p < 10^{-3}$ with 95% $CI = [0.0204; 0.0251]$) as shown in **Supplementary Material Table 4**, confirming the LCMV superiority in most of the analyses.

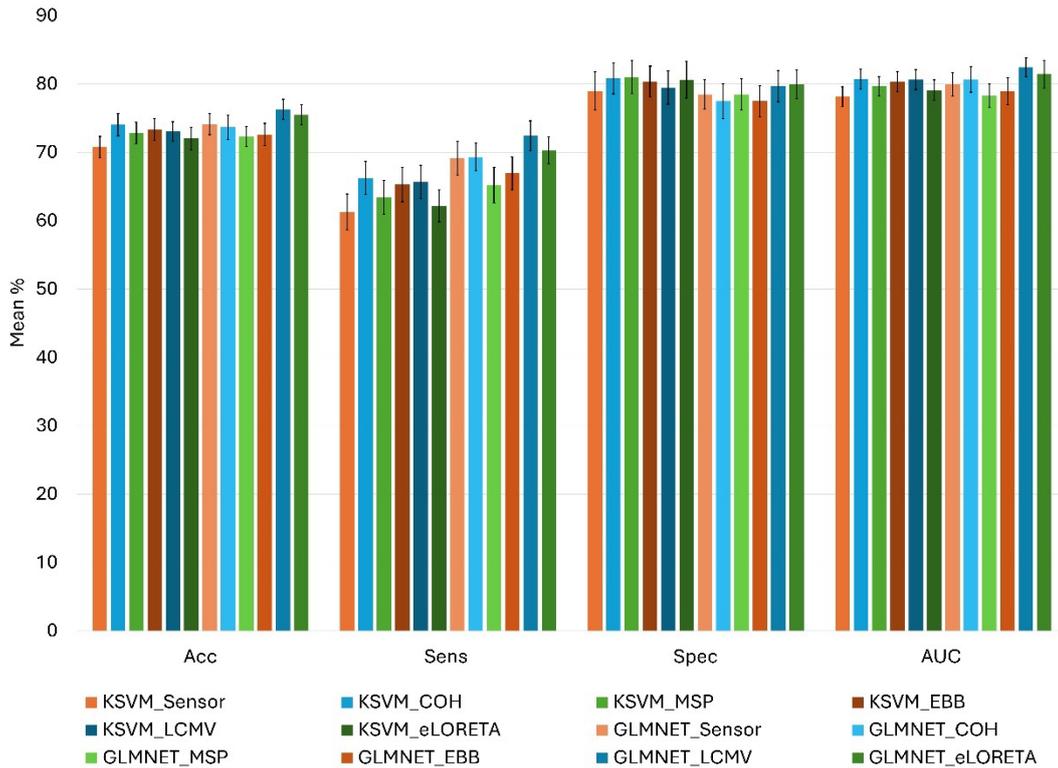

**Figure 6:** Comparative classification performance using KSVM and GLMNET models between outcomes for combining MEG's MAG and GRAD with MRI features, implemented separately for MEG sensor- and source-based analysis.

## 3.5 Feature selection and classification with GLMNET

Here, a more detailed revision is performed for the GLMNET outcomes to support the advantages of using this method for classification analysis. Firstly, for the exclusively MRI-based features analyses, **Fig. 7-8** present the results based on z-score-corrected features, which achieved an accuracy of 72.74% and AUC of 0.79 as reported above (**Section 3.3**), for the GLMNET coefficient estimates of mean ± error bars and derived signed mean z-score. **Fig. 7** provides a general picture of the sparsity of GLMNET solutions, with only a small number of relevant features. Complementarily, **Fig. 8** highlights the top 20 features selected by the GLMNET algorithm, including six MRI features identified as potential biomarkers for MCI or early AD progression. These features are the right parstriangularis, inferior lateral ventricle (Inf_Lat_Vent), left accumbens area, third ventricle, left lingual, and precentral regions. Remarkably, except for Inf_Lat_Vent, the contribution for the other selected features is mostly negative, which means that increments in the volume of these regions tend to diminish the predicted response. Considering than in the binary classification task with GLMNET, HC observations are marked as 0 while MCI are marked as 1, this means that large volumes observed during the biomarker measurement is an indication of healthiness. In contrast, for the Inf_Lat_Vent biomarker, its contribution its significant positive, meaning that a large volume observed in this biomarker measurement is an indication of cognitive decline.

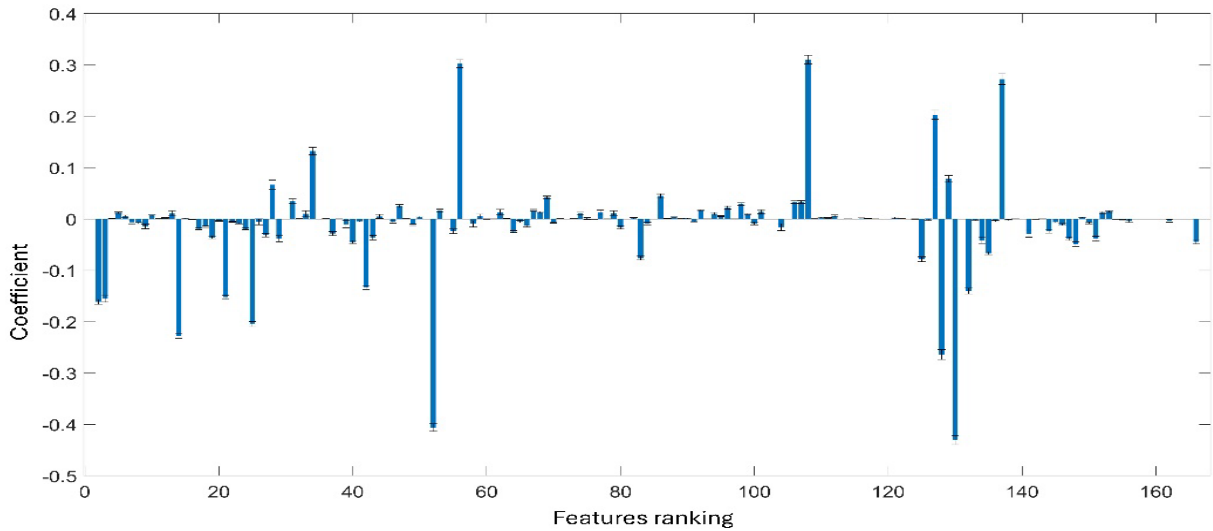

**Figure 7:** Mean bars ± error bars of estimated GLMNET coefficients across the 100 MC replications of the *K*=10 folds nested cross-validation procedure, shown for each of the 167 MRI features.

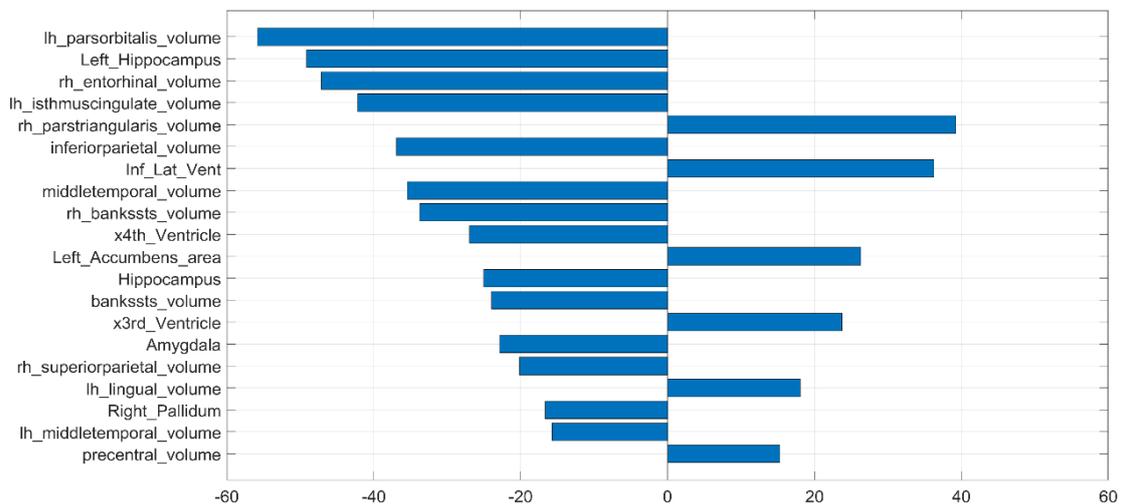

**Figure 8:** Signed z-score transform of estimated GLMNET coefficients across the 100 MC replications of the K=10 folds nested cross-validation procedure, shown for the top 20 more relevant MRI features.

Lastly, **Fig. 9** shows the results for the same calculations as above but obtained from the best classification performance of GLMNET based on MEG features (using LCMV inverse solution of GRAD signals), which achieved an accuracy of 72.94% and an AUC of 0.78 (see **Section 3.2**). As noticed, enhance interpretation can be achieved by projecting these results in the brain cortical map, separately by the analysed frequency bands. Specifically, if a brain region (R) shows larger positive z-score values for a particular frequency band (F), it strongly indicates that large positive changes observed in the measure of relative power spectrum at (R, F) favour the case prediction as MCI. Similar reasoning indicates that larger measured changes in regions and frequency bands with negative z-scores favour the case prediction as HC. Consistently with reports in the literature (Klimesch 1999; Vecchio et al. 2011), significant positive z-score values are prominently observed in the left-hemispheric temporal regions in theta band (highlighted with an overlayed orange circle), whereas significant negative z-score values are prominently observed in the alpha band somatosensory cortices (highlighted with an overlayed black rectangle).

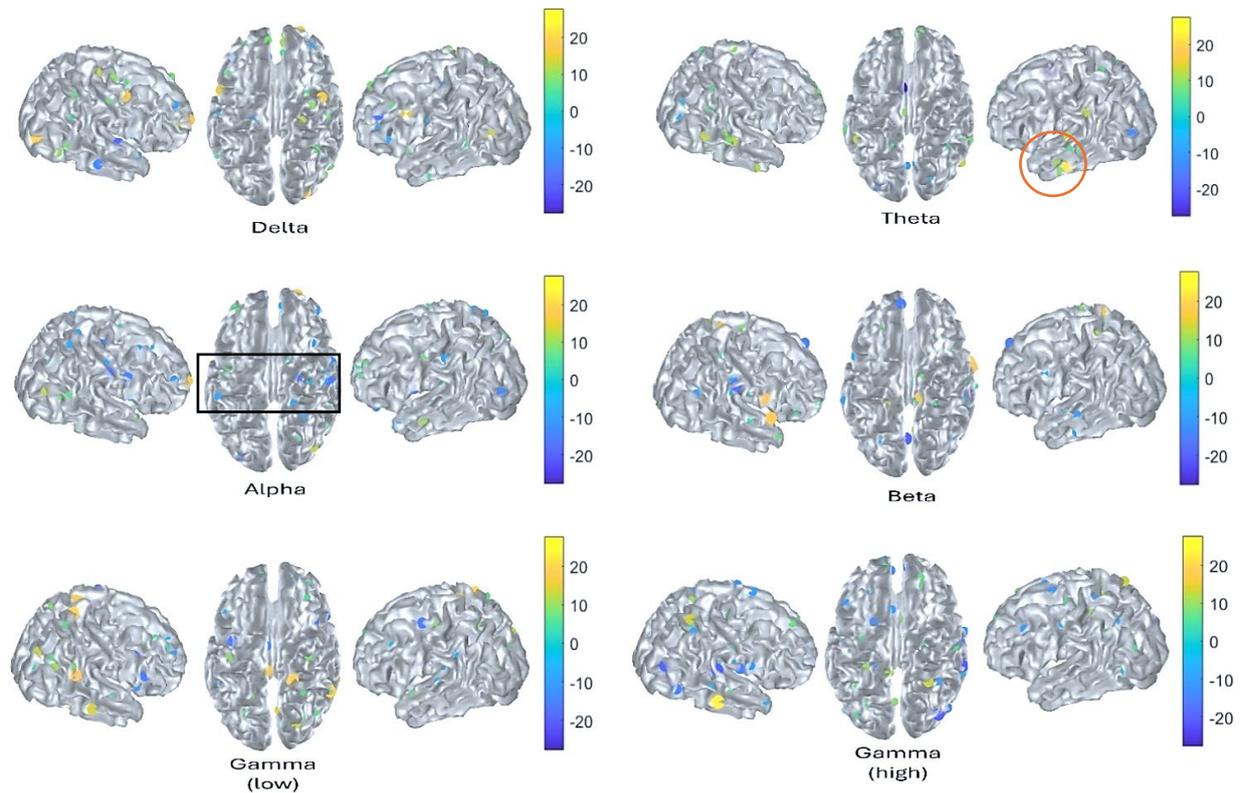

**Figure 9:** Signed z-score transform (color bars) of estimated GLMNET coefficients across the 100 MC replications of the K=10 folds nested cross-validation procedure, calculated similarly as in **Fig. 8** but for the MEG features (Total: 500 dipoles x 6 frequency bands), as shown projected in cortical maps for tree different views (adjacent left-hemisphere, top and right-hemisphere views) separately for the six frequency bands.

## 4. Discussion

In this study, we proposed the comparison and combination among performances obtained using different MEG and MRI pipelines to explore prospectively the better classification methods and feature combinations. Among these pipelines we compared the effectiveness of using MEG sensor- and source-based features, where for the latter we also compared the performance with respect to the utilization of different source localization methods, such as eLORETA and LCMV (Jaiswal et al. 2020a; Westner et al. 2022) . Distinctively, in comparison to other studies, we also compared the performance separately for the pipelines based on MEG's magnetometer (MAG) and gradiometers (GRAD) signals following the logic that they have different sensitives to noise sources as well as brain sources localized nearby on the cortical surface or deep in subcortical regions (Proudfoot et al. 2014; Hämäläinen et al. 1993). Ultimately, we also employed multiple classification approaches, as they can extract information from the features under different model assumptions. Particularly, we demonstrated the application of logistic regression combined with LASSO penalty, as implemented in the R's GLMNET package (Hastie, Qian, and Tay 2023; A. J. Friedman et al. 2010). In contrast to the other classifiers, where a priori selection of features was implemented based on the ReliefF method (Zhang, Li, and Chai 2022), the GLMNET tools enable the selection of features simultaneously with the gradual improvement on classification performance.

Our methodology was implemented based on different pipelines for pre-processing the MEG signals and extracting the MEG features in the sensor and source spaces, as well as for performing analogous operation for the MRI signals. The most critical was the classification pipeline, which implemented a Monte Carlo (MC) analysis involving 100 random replications of *K*=10 folds nested cross-validation, to robustly validate the evaluated classifiers and different features combination with holdout data generated using the nested strategy (see **Table 2**). The 100 generated scores for the different quality measures (Acc – accuracy, Sens – sensitivity, Spec – specificity and AUC – area under the receiver operator curve), obtained from each classifier performance based on the different feature combinations, were finally submitted to N-way ANOVA analysis to provide an assessment of the better approaches and prospective biomarkers.

## 4.1 Machine learning analysis

Here, we implemented a comprehensive analysis to achieve optimal performance in the selection of MEG based features and classifier models. With this purpose, we evaluated multiple classification models based on their performance in the literature (Delshad Vaghari, Ricardo Bruna, Laura E. Hughes, David Nesbitt, Roni Tibon, James B. Rowe, Fernando Maestu 2022; Van Hulse, Khoshgoftaar, and Napolitano 2007; Yue W, Wang Z, Chen H, Payne A 2018) and preliminary evaluation using the MATLAB Classification Leaner (MCL) app (**Table 3**). In previous studies, it has been reported that KSVM is one of the most preferred classification methods, including those utilizing the BioFIND dataset (Delshad Vaghari, Ricardo Bruna, Laura E. Hughes, David Nesbitt, Roni Tibon, James B. Rowe, Fernando Maestu 2022). Other research have explored the use of SVM with multi-kernel learning (Hughes et al. 2019; Vaghari, Kabir, and Henson 2022). Moreover, for most of these methods, we used the ReliefF technique to select *a priori* the most promising features before fitting the models within the classification pipeline. However, by exploring the ReliefF approach, we realized that this technique seems to be very effective to extract the more relevant features (see **Fig. 2**). In contrast, with the aim to evaluate another attractive strategy, we also used that GLMNET tools to evaluate logistic regression with LASSO penalty, with the advantage that feature selection are performed interleaved with the progression in fitting the classification model. Remarkably, logistic regression also has an interpretation edge (**Figs. 7-9**) over most of the other classification models. Overall, for the different evaluation using MEG and MRI features, we identified that KSVM was the best classifier when using only the MEG's MAG features, while GLMNET performed better when using only MEG's GRAD features and for all the combinations between MEG and MRI features.

## 4.2 MEG sensor and source localization analysis

There is an open debate in the literature concerning which inverse solution methods are more suitable for the accurate localization of brain cognitive functions, and even there is not clear evidence that demonstrate the superiority of any of these approaches or demonstrated high performance in brain source localization tasks. Therefore, sensor analyses are still very popular, despite of being considered inferior to source localization as in principle sensor signals are heavily affected by issues such as volume conduction and the reference problem. However, as demonstrated in the initial study involving the Biofind dataset (Hughes et al. 2019), sensor analysis can compete and even outperform source analysis.

To provide more evidence about this debate, we decided to select five different popular source localization methods, including COH (Baysian LORETA), EBB (Empirical Bayes Beamformer), MSP (Multiple Sparse Priors), eLORETA (exact Low Resolution Tomography Analysis), and LCMV (Linearly Constrained Minimum Variance Beamformer), as implemented in state-of-the-art brain neuroimaging toolboxes (Jaiswal et al. 2020a; Henson et al. 2019; Westner et al. 2022). As these methods follow different assumptions, our analysis provided a richer variety of features extracted from the application of these methods and, thereby, it makes more meaningful the utilization of the different extracted features, under different assumption, in the following classification analysis for a better assessment of the different evaluates methodologies.

**4.3 BioFIND dataset**

BioFIND is the most recent dataset to integrate MEG and MRI data for AD research, featuring multi-site data collection. Given its novelty, there has been limited publications associated with it. Therefore, one of our goals was also to improve the benchmark achieved for this dataset. With the use of multi-kernel learning with SVM, Vaghari et al. achieved an accuracy of 68% for features derived only from MEG at the sensor level, 65.2% from source level and 71% for features obtained exclusively through voxel-based morphometry (VBM) analysis, which is based on the MRI features (Vaghari, Kabir, and Henson 2022). Our findings reveal clearly superior performance, mainly for source-based MEG features obtain with the LCMV source localization approach, achieving an accuracy of 74.77%. For sensor space analysis of MEG-only based features, our results were more similar to the previous study with an accuracy of 69.28%. For the MRI analysis, in contrast to VBM we used Freesurfer to extract the brain anatomical features, achieving a modest improved accuracy of 72.74% using GLMNET with z-score-corrected data.

To the best of our knowledge, our current work is the first publication involving the BioFIND dataset which proposes the separate analysis of MEG's MAG and GRAD signals, and their combination with MRI signals, as well as the use of multiple source localization methods. We implemented separate classification analyses for the MEG's MAG and GRAD signals to avoid bias in the MEG inverse solution due to the different noise and signal scales in these sensors (Dash et al. 2021; Garcés et al. 2017a; Gross et al. 2013). Our initial expectancy was that results should be better for features extracted by using GRAD signals, primarily due to their superior noise suppression capabilities (Dash et al. 2021; Garcés et al. 2017a). However, our study showed the highest performance, using single modality data, for the features extracted from the MAG signal analyses, with an accuracy of 74.77% compared to an accuracy of 73.24% when utilizing the GRAD features. Moreover, a comparison between uncorrected and a data-correction procedure for MEG-based features revealed that the use of the data correction procedure proposed in the first BioFIND paper was ineffective ($F_{(1,16632)} = 1695.7$ with $p < 10^{-7}$), consistently with their study (Delshad Vaghari, Ricardo Bruna, Laura E. Hughes, David Nesbitt, Roni Tibon, James B. Rowe, Fernando Maestu 2022). This negative result about data correction may be attributed to the loss of critical information due to the harmonization procedure. Notably, there is a significant difference in the proportion of MCI and HC between the participants recorded in the Spain and UK populations ($F_{(1,322)} = 4.53$ with $p = 0.034$). Therefore, using the covariate correction approach proposed in the first BioFIND paper (Hughes et al. 2019), which involves the variable "site" as a covariate, it may indeed cause the loss of some critical information in the features that can be actually attributed to real HC vs MCI differences.

Moreover, we explored a potentially superior classification technique with respect to previous analyses. By using GLMNET (Kang et al. 2019), we were able to identify affected brain regions in MCI subjects that can be used as biomarkers for early detection of AD (**Figs. 8-9**). Anatomical biomarkers detected by GLMNET included brain regions that may be affected by early AD progression (Sun et al. 2018; de Vos et al. 2016; Schroeter et al. 2009; Poloni et al. 2021; Ledig et al. 2018), such as hippocampus, entorhinal, parstriangularis, inferior lateral ventricle, among other regions (**Fig. 8**). Additionally, in our analysis of MEG features, GLMNET also provided us with insightful results about the topographic spectral characteristic of extracted features (**Fig. 9**). An interesting observation from our results is that significant positive theta-band features are prominently observed in both the left and right hemispheres. As HC/MCI classes are labelled as 0/1, respectively, in our GLMNET analysis, it means that larger than average theta activity for a particular case, could be associated with increased risk of cognitive decline. In contrast, significant negative alpha-band features were observed in the somatosensory cortices, meaning that for a particular subject's MEG data, larger than average alpha activity in these regions could be associated with brain healthiness.

## 5. Limitation

Despite we tested several inverse solutions, this investigation is far from best as the debate over the most appropriate inverse solution approaches is still wide open (Pascual-Marqui 2002; Soufflet and Boeijinga 2005; Grech et al. 2008; Michel et al. 2004; S. Baillet, Mosher, and Leahy 2001; Liu, Dale, and Belliveau 2002; Hämäläinen et al. 1993). Moreover, although in principle functional connectivity analysis should produce better results it is hard to say when they are based on a maybe suboptimal inverse solution (Mahjoory et al. 2017; Hincapié et al. 2017). Future investigation should address these two concerns: whether improved source localization and functional connectivity methods can contribute the advancement of AD biomarker research based on MEG features. On the other hand, in some analysis we faced a large number of features with high correlations among them. This is a critical issue that can affect the classification performance. Using simultaneous feature selection and classification with GLMNET could help to address these issues. However, in our study, we explored the most promising candidates of GLMNET tools. For example, it has been proposed that that using $\alpha < 1$ in **Eq. (2)**, which implies using both LASSO (L1 norm) and L2-norm penalty functions can bring more stability to classification with GLMNET models. Similarly, the GLMNET tool enables the use of nonpositive and/or nonnegative constraints. Knowing in advance that cognitive decline is associated with larger than average brain atrophy than in healthy aging, and similarly for the enlargement of ventricular volumes, these sign constraints could be considered to reflect this knowledge.

## 6. Conclusion

Despite the availability of multiple options, e.g., among different inverse solution and classification methods, we provided a comprehensive approach that combines MEG and MRI features. The use of Monte-Carlo analysis consisting of 100 replications of a 10-fold nested cross-validation procedure was critically important to realise the danger of overfitting and evaluate the "better" techniques. Even when some of the explored classifiers implement different hyperparameter evaluation strategies, we observed that overfitting can appear even in very tightly controlled scenarios, and therefore it is essential to validate the outcome with unseen data, as implemented in our study through nested cross-validation. The classifiers that shown superior performance are identified as KSVM and GLMNET. Particularly, the GLMNET approach deserved increased attention in the study of AD progression with MEG and MRI features due to enhanced interpretation of the results. Finally, it is shown that exclusively using only MEG features can contribute to comparable or higher performance than using strictly only MRI features, with the LCMV and eLORETA inverse solutions allowing for better performance among compared approaches; however, as also demonstrated, combination of source-based uncorrected MEG and z-score-corrected MRI features is preferable.

### Ethics statement of possible datasets to be used in this study.

No data was collected during the implementation of this study. Data not publicly available due to privacy or ethical restrictions. Applications for access should be submitted to the data access committee, which can be reached through: https://www.dementiasplatform.uk/research-hub/data-portal/featured-cohort-biofind.

### Declaration of Competing Interest

The authors declare that they do not possess any discernible competing financial interests or personal affiliations that might have conceivably impacted the research presented in this manuscript.

### Author Contribution

**Alwani Liyana Ahamd:** Conceptualization, Methodology, Software, Formal analysis, Writing- original draft, Writing- review & editing; **Jose Sanchez-Bornot:** Methodology, Software, Writing- review & editing; **Roberto C. Sotero:** Writing- review & editing; **Damien Coyle:** Writing- review & editing, **Zamzuri Idris**: Writing- review & editing, **Ibrahima Faye:** Methodology, Writing- review & editing, Funding acquisition

**Acknowledgement**

The authors extend their gratitude to all BioFIND and MRC Dementias Platform UK (DPUK) members for their support. Special thanks are due to Catrin Morris for her continuous assistance, and also to Elen Golightly, Fatemeh Torabi, Jason W. Price, and Kevin J. Shaw for their technical support.

This work was conducted using the resources of the DPUK, which is a Public-Private Partnership funded by the Medical Research Council (MR/L023784/1 and MR/009076/1). For further information on this resource visit www.dementiasplatform.uk. ALA would like to thank Universiti Sains Malaysia and Ministry of Higher Education (MOHE) Malaysia (Scholarship *Hadiah Latihan Persekutuan (HLP)*) for sponsoring and Universiti Teknologi PETRONAS for supporting this study. RCS was supported by Grant 222300868 from the Alberta Innovates LevMax program.

**References:**

Aggarwal, Srishty, and Supratim Ray. 2023. "Slope of the Power Spectral Density Flattens at Low Frequencies (<150 Hz) with Healthy Aging but Also Steepens at Higher Frequency (>200 Hz) in Human Electroencephalogram." *Cerebral Cortex Communications* 4 (2): 1–12. https://doi.org/10.1093/texcom/tgad011.

Ahmad, Alwani Liyana, Jose Sanchez-Bornot, Roberto C. Sotero, Damien Coyle, Zamzuri Idris, and Ibrahima Faye. 2024. "A Machine Learning Approach for Identifying Anatomical Biomarkers of Early Mild Cognitive Impairment." *ArXiv Preprint*, May. http://arxiv.org/abs/2407.00040.

Asadzadeh, Shiva, Tohid Yousefi Rezaii, Soosan Beheshti, Azra Delpak, and Saeed Meshgini. 2020. "A Systematic Review of EEG Source Localization Techniques and Their Applications on Diagnosis of Brain Abnormalities." *Journal of Neuroscience Methods* 339 (April). https://doi.org/10.1016/j.jneumeth.2020.108740.

Baillet, S., J.C. Mosher, and R.M. Leahy. 2001. "Electromagnetic Brain Mapping." *IEEE Signal Processing Magazine* 18 (6): 14–30. https://doi.org/10.1109/79.962275.

Baillet, Sylvain. 2017. "Magnetoencephalography for Brain Electrophysiology and Imaging" 20 (3). https://doi.org/10.1038/nn.4504.

Bénar, Christian George, Jayabal Velmurugan, Victor J. López-Madrona, Francesca Pizzo, and Jean Michel Badier. 2021. "Detection and Localization of Deep Sources in Magnetoencephalography: A Review." *Current Opinion in Biomedical Engineering* 18 (March): 100285. https://doi.org/10.1016/j.cobme.2021.100285.

Bruce Fischl, David H. Salat, André J.W. van der Kouwe, Nikos Makris, Florent Ségonne, Brian T. Quinn, Anders M. Dale. 2004. "Sequence-Independent Segmentation of Magnetic Resonance Images." *NeuroImage* 23 (SUPPL. 1): 69–84. https://doi.org/10.1016/j.neuroimage.2004.07.016.

Bruña, Ricardo, David López-Sanz, Fernando Maestú, Ann D. Cohen, Anto Bagic, Ted Huppert, Tae Kim, Rebecca E. Roush, Betz Snitz, and James T. Becker. 2023. "MEG Oscillatory Slowing in Cognitive Impairment Is Associated with the Presence of Subjective Cognitive Decline." *Clinical EEG and Neuroscience* 54 (1): 73–81. https://doi.org/10.1177/15500594221072708.

Dafflon, Jessica, Walter H.L. Pinaya, Federico Turkheimer, James H. Cole, Robert Leech, Mathew A.

Harris, Simon R. Cox, Heather C. Whalley, Andrew M. McIntosh, and Peter J. Hellyer. 2020. "An Automated Machine Learning Approach to Predict Brain Age from Cortical Anatomical Measures." *Human Brain Mapping* 41 (13): 3555–66. https://doi.org/10.1002/hbm.25028.

Darvas, F., D. Pantazis, E. Kucukaltun-Yildirim, and R. M. Leahy. 2004. "Mapping Human Brain Function with MEG and EEG: Methods and Validation." *NeuroImage* 23 (SUPPL. 1): 289–99. https://doi.org/10.1016/j.neuroimage.2004.07.014.

Dash, Debadatta, Paul Ferrari, Abbas Babajani-Feremi, Amir Borna, Peter D.D. Schwindt, and Jun Wang. 2021. "Magnetometers vs Gradiometers for Neural Speech Decoding." *Proceedings of the Annual International Conference of the IEEE Engineering in Medicine and Biology Society, EMBS*, 6543–46. https://doi.org/10.1109/EMBC46164.2021.9630489.

Davenport, Elizabeth M. 2023. "POSTER PRESENTATION Magnetoencephalography ( MEG ) Spectral Signature of Cognitive Impairment - A Perspective on Pathophysiological Changes with Blood Pressure" 19: 1–2. https://doi.org/10.1002/alz.065710.

Delshad Vaghari, Ricardo Bruna, Laura E. Hughes, David Nesbitt, Roni Tibon, James B. Rowe, Fernando Maestu, Richard N. Henson. 2022. "A Multi-Site, Multi-Participant Magnetoencephalography Resting-State Dataset to Study Dementia: The BioFIND Dataset." *NeuroImage* 258 (May): 119344. https://doi.org/10.1016/j.neuroimage.2022.119344.

Fergus, P., A. Hussain, David Hignett, D. Al-Jumeily, Khaled Abdel-Aziz, and Hani Hamdan. 2016. "A Machine Learning System for Automated Whole-Brain Seizure Detection." *Applied Computing and Informatics* 12 (1): 70–89. https://doi.org/10.1016/j.aci.2015.01.001.

Friedman, Author Jerome, Trevor Hastie, Rob Tibshirani, and Maintainer Trevor Hastie. 2010. "Package ' Glmnet .'"

Friedman, Jerome, Trevor Hastie, and Rob Tibshirani. 2010. "Regularization Paths for Generalized Linear Models via Coordinate Descent." *Journal of Statistical Software* 33 (1): 1–22. https://doi.org/10.18637/jss.v033.i01.

Garcés, Pilar, David López-Sanz, Fernando Maestú, and Ernesto Pereda. 2017a. "Choice of Magnetometers and Gradiometers after Signal Space Separation." *Sensors (Switzerland)* 17 (12): 1–13. https://doi.org/10.3390/s17122926.

———. 2017b. "Choice of Magnetometers and Gradiometers after Signal Space Separation." *Sensors* 17 (12): 2926. https://doi.org/10.3390/s17122926.

Grech, Roberta, Tracey Cassar, Joseph Muscat, Kenneth P. Camilleri, Simon G. Fabri, Michalis Zervakis, Petros Xanthopoulos, Vangelis Sakkalis, and Bart Vanrumste. 2008. "Review on Solving the Inverse Problem in EEG Source Analysis." *Journal of NeuroEngineering and Rehabilitation* 5: 1–33. https://doi.org/10.1186/1743-0003-5-25.

Gross, Joachim, Sylvain Baillet, Gareth R. Barnes, Richard N. Henson, Arjan Hillebrand, Ole Jensen, Karim Jerbi, et al. 2013. "Good Practice for Conducting and Reporting MEG Research." *NeuroImage* 65: 349–63. https://doi.org/10.1016/j.neuroimage.2012.10.001.

Hämäläinen, Matti, Riitta Hari, Risto J. Ilmoniemi, Jukka Knuutila, and Olli V. Lounasmaa. 1993. "Magnetoencephalography Theory, Instrumentation, and Applications to Noninvasive Studies of the Working Human Brain." *Reviews of Modern Physics* 65 (2): 413–97. https://doi.org/10.1103/RevModPhys.65.413.

Hämäläinen, Matti, Mingxiong Huang, and Susan M. Bowyer. 2020. "Magnetoencephalography Signal Processing, Forward Modeling, Magnetoencephalography Inverse Source Imaging, and Coherence Analysis." *Neuroimaging Clinics of North America* 30 (2): 125–43.

https://doi.org/10.1016/j.nic.2020.02.001.

Hastie, Trevor, Junyang Qian, and Kenneth Tay. 2023. "An Introduction to Glmnet." *CRAN R Repositary*, 1–38.

Henson, Richard N, Hunar Abdulrahman, Guillaume Flandin, and Vladimir Litvak. 2019. "Multimodal Integration of M / EEG and f / MRI Data in SPM12" 13 (April): 1–22. https://doi.org/10.3389/fnins.2019.00300.

Hincapié, Ana Sofía, Jan Kujala, Jérémie Mattout, Annalisa Pascarella, Sebastien Daligault, Claude Delpuech, Domingo Mery, Diego Cosmelli, and Karim Jerbi. 2017. "The Impact of MEG Source Reconstruction Method on Source-Space Connectivity Estimation: A Comparison between Minimum-Norm Solution and Beamforming." *NeuroImage* 156 (May): 29–42. https://doi.org/10.1016/j.neuroimage.2017.04.038.

Hughes, Laura E., Richard N. Henson, Ernesto Pereda, Ricardo Bruña, David López-Sanz, Andrew J. Quinn, Mark W. Woolrich, Anna C. Nobre, James B. Rowe, and Fernando Maestú. 2019. "Biomagnetic Biomarkers for Dementia: A Pilot Multicentre Study with a Recommended Methodological Framework for Magnetoencephalography." *Alzheimer's and Dementia: Diagnosis, Assessment and Disease Monitoring* 11: 450–62. https://doi.org/10.1016/j.dadm.2019.04.009.

Hulse, Jason Van, Taghi M. Khoshgoftaar, and Amri Napolitano. 2007. "Experimental Perspectives on Learning from Imbalanced Data." *ACM International Conference Proceeding Series* 227: 935–42. https://doi.org/10.1145/1273496.1273614.

Jaiswal, Amit, Jukka Nenonen, Matti Stenroos, Alexandre Gramfort, Sarang S. Dalal, Britta U. Westner, Vladimir Litvak, et al. 2020a. "Comparison of Beamformer Implementations for MEG Source Localization." *NeuroImage* 216 (March): 116797. https://doi.org/10.1016/j.neuroimage.2020.116797.

Jaiswal, Amit, Jukka Nenonen, Matti Stenroos, Alexandre Gramfort, Sarang S Dalal, Britta U Westner, Vladimir Litvak, et al. 2020b. "Comparison of Beamformer Implementations for MEG Source Localization." *NeuroImage* 216 (February): 116797. https://doi.org/https://doi.org/10.1016/j.neuroimage.2020.116797.

Kabir, Md Humaun, Shabbir Mahmood, Abdullah Al Shiam, Abu Saleh Musa Miah, Jungpil Shin, and Md Khademul Islam Molla. 2023. "Investigating Feature Selection Techniques to Enhance the Performance of EEG-Based Motor Imagery Tasks Classification." *Mathematics* 11 (8): 1–19. https://doi.org/10.3390/math11081921.

Kang, Chuanze, Yanhao Huo, Lihui Xin, Baoguang Tian, and Bin Yu. 2019. "Feature Selection and Tumor Classification for Microarray Data Using Relaxed Lasso and Generalized Multi-Class Support Vector Machine." *Journal of Theoretical Biology* 463: 77–91. https://doi.org/10.1016/j.jtbi.2018.12.010.

Kira, Kenji, Kenji Kira, Larry A Rendell, and Larry A Rendell. 1992. "The Feature Selection Problem: Traditional Methods and a New Algorithm." https://scholar.google.com/scholar?q=The feature selection problem: traditional methods and a new algorithm.

Klimesch, Wolfgang. 1999. "EEG Alpha and Theta Oscillations Reflect Cognitive and Memory Performance: A RKlimesch, W. (1999). EEG Alpha and Theta Oscillations Reflect Cognitive and Memory Performance: A Review and Analysis. Brain Research Reviews, 29(2-3), 169–195. Doi:10.1016/S016." *Brain Research Reviews* 29 (2–3): 169–95. https://doi.org/10.1016/S0165-0173(98)00056-3.

Koikkalainen, Juha, Harri Pölönen, Jussi Mattila, Mark van Gils, Hilkka Soininen, and Jyrki Lötjönen. 2012. "Improved Classification of Alzheimer's Disease Data via Removal of Nuisance Variability." *PLoS ONE* 7 (2). https://doi.org/10.1371/journal.pone.0031112.

Koponen, Lari M., and Angel V. Peterchev. 2020. "Transcranial Magnetic Stimulation: Principles and Applications." In *Neural Engineering*, 245–70. Cham: Springer International Publishing. https://doi.org/10.1007/978-3-030-43395-6_7.

Lancaster, Jack L., Marty G. Woldorff, Lawrence M. Parsons, Mario Liotti, Catarina S. Freitas, Lacy Rainey, Peter V. Kochunov, Dan Nickerson, Shawn A. Mikiten, and Peter T. Fox. 2000. "Automated Talairach Atlas Labels for Functional Brain Mapping." *Human Brain Mapping* 10 (3): 120–31. https://doi.org/10.1002/1097-0193(200007)10:3<120::AID-HBM30>3.0.CO;2-8.

Ledig, Christian, Andreas Schuh, Ricardo Guerrero, Rolf A. Heckemann, and Daniel Rueckert. 2018. "Structural Brain Imaging in Alzheimer's Disease and Mild Cognitive Impairment: Biomarker Analysis and Shared Morphometry Database." *Scientific Reports* 8 (1): 1–16. https://doi.org/10.1038/s41598-018-29295-9.

Liu, Arthur K., Anders M. Dale, and John W. Belliveau. 2002. "Monte Carlo Simulation Studies of EEG and MEG Localization Accuracy." *Human Brain Mapping* 16 (1): 47–62. https://doi.org/10.1002/hbm.10024.

López-sanz, David, Ricardo Bruña, María Luisa Delgado-losada, Ramón López-higes, Alberto Marcos-dolado, Fernando Maestú, and Stefan Walter. 2019. "Electrophysiological Brain Signatures for the Classification of Subjective Cognitive Decline : Towards an Individual Detection in the Preclinical Stages of Dementia" 3: 1–10.

Lv, Zhihan, Liang Qiao, Qingjun Wang, and Francesco Piccialli. 2021. "Advanced Machine-Learning Methods for Brain-Computer Interfacing." *IEEE/ACM Transactions on Computational Biology and Bioinformatics* 18 (5): 1688–98. https://doi.org/10.1109/TCBB.2020.3010014.

Ma D, Popuri K, Bhalla M, Sangha O, Lu D, Cao J, Jacova C, Wang L, Beg MF; Alzheimer's Disease Neuroimaging Initiative. 2019. "Quantitative Assessment of Field Strength, Total Intracranial Volume, Sex, and Age Effects on the Goodness of Harmonization for Volumetric Analysis on the ADNI Database." *Human Brain Mapping* 40 (5): 1507–27. https://doi.org/10.1002/hbm.24463.

Mahjoory, Keyvan, Vadim V. Nikulin, Loïc Botrel, Klaus Linkenkaer-Hansen, Marco M. Fato, and Stefan Haufe. 2017. "Consistency of EEG Source Localization and Connectivity Estimates." *NeuroImage* 152 (May): 590–601. https://doi.org/10.1016/J.NEUROIMAGE.2017.02.076.

Mandal, Pravat K, Anwesha Banerjee, Manjari Tripathi, and Ankita Sharma. 2018. "A Comprehensive Review of Magnetoencephalography ( MEG ) Studies for Brain Functionality in Healthy Aging and Alzheimer ' s Disease ( AD )" 12 (August). https://doi.org/10.3389/fncom.2018.00060.

Michel, Christoph M., Micah M. Murray, Göran Lantz, Sara Gonzalez, Laurent Spinelli, and Rolando Grave de Peralta. 2004. "EEG Source Imaging." *Clinical Neurophysiology* 115 (10): 2195–2222. https://doi.org/10.1016/j.clinph.2004.06.001.

Molla, Md Khademul Islam, Abdullah Al Shiam, Md Rabiul Islam, and Toshihisa Tanaka. 2020. "Discriminative Feature Selection-Based Motor Imagery Classification Using EEG Signal." *IEEE Access* 8: 98255–65. https://doi.org/10.1109/ACCESS.2020.2996685.

Mosher, John C., Richard M. Leahy, and Paul S. Lewis. 1999. "EEG and MEG: Forward Solutions for Inverse Methods." *IEEE Transactions on Biomedical Engineering* 46 (3): 245–59. https://doi.org/10.1109/10.748978.

Pantazis, Dimitrios, and Richard M. Leahy. 2006. "Imaging the Human Brain with Magnetoencephalography." In *Handbook of Research on Informatics in Healthcare and Biomedicine*, 294–302. IGI Global. https://doi.org/10.4018/978-1-59140-982-3.ch038.

Parvandeh, Saeid, Hung Wen Yeh, Martin P. Paulus, and Brett A. McKinney. 2020. "Consensus Features Nested Cross-Validation." *Bioinformatics* 36 (10): 3093–98. https://doi.org/10.1093/bioinformatics/btaa046.

Pascual-Marqui, R D. 2002. "Standardized Low-Resolution Brain Electromagnetic Tomography (SLORETA): Technical Details." *Methods and Findings in Experimental and Clinical Pharmacology* 24 Suppl D: 5–12. http://www.ncbi.nlm.nih.gov/pubmed/12575463.

Poloni, Katia M., Italo A. Duarte de Oliveira, Roger Tam, and Ricardo J. Ferrari. 2021. "Brain MR Image Classification for Alzheimer's Disease Diagnosis Using Structural Hippocampal Asymmetrical Attributes from Directional 3-D Log-Gabor Filter Responses." *Neurocomputing* 419: 126–35. https://doi.org/10.1016/j.neucom.2020.07.102.

Popuri, Karteek, Da Ma, Lei Wang, and Mirza Faisal Beg. 2020. "Using Machine Learning to Quantify Structural MRI Neurodegeneration Patterns of Alzheimer's Disease into Dementia Score: Independent Validation on 8,834 Images from ADNI, AIBL, OASIS, and MIRIAD Databases." *Human Brain Mapping* 41 (14): 4127–47. https://doi.org/10.1002/hbm.25115.

Poza, Jesus, Roberto Hornero, Daniel Abasolo, Alberto Fernandez, and Javier Escudero. 2007. "Analysis of Spontaneous MEG Activity in Patients with Alzheimer's Disease Using Spectral Entropies." In *2007 29th Annual International Conference of the IEEE Engineering in Medicine and Biology Society*, 6179–82. IEEE. https://doi.org/10.1109/IEMBS.2007.4353766.

Proudfoot, Malcolm, Mark W. Woolrich, Anna C. Nobre, and Martin R. Turner. 2014. "Magnetoencephalography." *Practical Neurology* 14 (5): 336–43. https://doi.org/10.1136/practneurol-2013-000768.

R. A. I. Bethlehem, J. Seidlitz, S. R. White, J. W. Vogel, K. M. Anderson, C. Adamson, S. Adler, G. S. Alexopoulos, E. Anagnostou, A. Areces-Gonzalez, D. E. Astle, B. Auyeung, M. Ayub, J. Bae, G. Ball, S. Baron-Cohen, R. Beare, S. A. Bedford, V. Benegal, E. T. Bullmore & A. F. Alexander-Bloch. 2022. "Brain Charts for the Human Lifespan." *Nature* 604 (February). https://doi.org/10.1038/s41586-022-04554-y.

Sanchez-Bornot, Jose M., Maria E. Lopez, Ricardo Bruña, Fernando Maestu, Vahab Youssofzadeh, Su Yang, David P. Finn, et al. 2021. "High-Dimensional Brain-Wide Functional Connectivity Mapping in Magnetoencephalography." *Journal of Neuroscience Methods* 348. https://doi.org/10.1016/j.jneumeth.2020.108991.

Sanchez-Bornot, Jose M., Roberto C. Sotero, Scott Kelso, and Damien Coyle. 2022. "Solving Large-Scale MEG/EEG Source Localization and Functional Connectivity Problems Simultaneously Using State-Space Models," no. 1. https://arxiv.org/abs/2208.12854v1.

Sanchez-Rodriguez, Lazaro M, and Yasser Iturria Medina. 2023. "Neural Mass Framework for Decoding Amyloid and Tau Impact on Neuronal Activity under Alzheimer's Disease." *Alzheimer's & Dementia* 19 (S3): 1–2. https://doi.org/10.1002/alz.064972.

Scheijbeler, Elliz P., Anne M. van Nifterick, Cornelis J. Stam, Arjan Hillebrand, Alida A. Gouw, and Willem de Haan. 2022. "Network-Level Permutation Entropy of Resting-State MEG Recordings: A Novel Biomarker for Early-Stage Alzheimer's Disease?" *Network Neuroscience* 6 (2): 382–400. https://doi.org/10.1162/netn_a_00224.

Schoffelen, Jan Mathijs, and Joachim Gross. 2009. "Source Connectivity Analysis with MEG and EEG." *Human Brain Mapping* 30 (6): 1857–65. https://doi.org/10.1002/hbm.20745.

Schroeter, Matthias L., Timo Stein, Nina Maslowski, and Jane Neumann. 2009. "Neural Correlates of Alzheimer's Disease and Mild Cognitive Impairment: A Systematic and Quantitative Meta-Analysis Involving 1351 Patients." *NeuroImage* 47 (4): 1196–1206. https://doi.org/10.1016/j.neuroimage.2009.05.037.

Ségonne, F., A. M. Dale, E. Busa, M. Glessner, D. Salat, H. K. Hahn, and B. Fischl. 2004. "A Hybrid Approach to the Skull Stripping Problem in MRI." *NeuroImage* 22 (3): 1060–75. https://doi.org/10.1016/j.neuroimage.2004.03.032.

Singh, Mahender Kumar, and Krishna Kumar Singh. 2021. "A Review of Publicly Available Automatic Brain Segmentation Methodologies, Machine Learning Models, Recent Advancements, and Their Comparison." *Annals of Neurosciences* 28 (1–2): 82–93. https://doi.org/10.1177/0972753121990175.

Soufflet, Laurent, and Peter H. Boeijinga. 2005. "Linear Inverse Solutions: Simulations from a Realistic Head Model in MEG." *Brain Topography* 18 (2): 87–99. https://doi.org/10.1007/s10548-005-0278-6.

Sun, Zhuo, Yuchuan Qiao, Boudewijn P.F. Lelieveldt, and Marius Staring. 2018. "Integrating Spatial-Anatomical Regularization and Structure Sparsity into SVM: Improving Interpretation of Alzheimer's Disease Classification." *NeuroImage* 178 (February 2018): 445–60. https://doi.org/10.1016/j.neuroimage.2018.05.051.

Tagliazucchi, Enzo, and Helmut Laufs. 2015. "Multimodal Imaging of Dynamic Functional Connectivity." *Frontiers in Neurology* 6 (FEB): 1–9. https://doi.org/10.3389/fneur.2015.00010.

Tandel, Gopal S., Mainak Biswas, Omprakash G. Kakde, Ashish Tiwari, Harman S. Suri, Monica Turk, John R. Laird, et al. 2019. "A Review on a Deep Learning Perspective in Brain Cancer Classification." *Cancers* 11 (1). https://doi.org/10.3390/cancers11010111.

Tay, J. Kenneth, Balasubramanian Narasimhan, and Trevor Hastie. 2023. "Elastic Net Regularization Paths for All Generalized Linear Models." *Journal of Statistical Software* 106 (1). https://doi.org/10.18637/jss.v106.i01.

Vaghari, Delshad, Ehsanollah Kabir, and Richard N. Henson. 2022. "Late Combination Shows That MEG Adds to MRI in Classifying MCI versus Controls." *NeuroImage* 252 (February). https://doi.org/10.1016/j.neuroimage.2022.119054.

Vallarino, Elisabetta, Ana Sofia Hincapié, Karim Jerbi, Richard M. Leahy, Annalisa Pascarella, Alberto Sorrentino, and Sara Sommariva. 2023. "Tuning Minimum-Norm Regularization Parameters for Optimal MEG Connectivity Estimation." *NeuroImage* 281 (September): 120356. https://doi.org/10.1016/j.neuroimage.2023.120356.

Vecchio, Fabrizio, Roberta Lizio, Giovanni B. Frisoni, Raffaele Ferri, Guido Rodriguez, and Claudio Babiloni. 2011. "Electroencephalographic Rhythms in Alzheimer's Disease." *International Journal of Alzheimer's Disease* 2011. https://doi.org/10.4061/2011/927573.

Velmurugan, Jayabal, Sanjib Sinha, and Parthasarathy Satishchandra. 2014. "Magnetoencephalography Recording and Analysis." *Annals of Indian Academy of Neurology* 17 (SUPPL. 1). https://doi.org/10.4103/0972-2327.128678.

Verdoorn, Todd A., J. Riley McCarten, David B. Arciniegas, Richard Golden, Leslie Moldauer, Apostolos Georgopoulos, Scott Lewis, et al. 2011. "Evaluation and Tracking of Alzheimer's Disease Severity Using Resting-State Magnetoencephalography." Edited by J. Wesson Ashford, Allyson Rosen, Maheen Adamson, Peter Bayley, Osama Sabri, Ansgar Furst, Sandra E. Black, and Michael Weiner. *Journal of Alzheimer's Disease* 26 (s3): 239–55. https://doi.org/10.3233/JAD-

2011-0056.

Verma, Parul, Hannah Lerner, and Danielle Mizuiri. 2023. "Impaired Long-Range Excitatory Time Scale Predicts Abnormal Neural Oscillations and Cognitive de Cits in Alzheimer ' s Disease." *Research Square* 3. https://doi.org/10.21203/rs.3.rs-2579392/v3.

Vos, Frank de, Tijn M. Schouten, Anne Hafkemeijer, Elise G.P. Dopper, John C. van Swieten, Mark de Rooij, Jeroen van der Grond, and Serge A.R.B. Rombouts. 2016. "Combining Multiple Anatomical MRI Measures Improves Alzheimer's Disease Classification." *Human Brain Mapping* 37 (5): 1920–29. https://doi.org/10.1002/hbm.23147.

Wang, R., Wang, J., Yu, H. et al. 2015. "Power Spectral Density and Coherence Analysis of Alzheimer's EEG." *Cognitive Neurodynamics* 9: 291–304. https://doi.org/https://doi.org/10.1007/s11571-014-9325-x.

Westner, Britta U., Sarang S. Dalal, Alexandre Gramfort, Vladimir Litvak, John C. Mosher, Robert Oostenveld, and Jan Mathijs Schoffelen. 2022. "A Unified View on Beamformers for M/EEG Source Reconstruction." *NeuroImage* 246 (June 2021): 118789. https://doi.org/10.1016/j.neuroimage.2021.118789.

William D. Penny, Karl J. Friston, John T. Ashburner, Stefan J. Kiebel, Thomas E. Nichols, ed. 2011. *Statistical Parametric Mapping: The Analysis of Functional Brain Images*. Elsevier.

Yang, Su, Jose Miguel Sanchez Bornot, Kongfatt Wong-Lin, and Girijesh Prasad. 2019. "M/EEG-Based Bio-Markers to Predict the MCI and Alzheimer's Disease: A Review from the ML Perspective." *IEEE Transactions on Biomedical Engineering* 66 (10): 2924–35. https://doi.org/10.1109/TBME.2019.2898871.

Yook, Yeeun. 2022. "The Study of Neuronal Hyperactivity in the Early Stage of Alzheimer's Disease." *Alzheimer's and Dementia* 18 (S4): 61398. https://doi.org/10.1002/alz.061398.

Yue W, Wang Z, Chen H, Payne A, Liu X. 2018. "Machine Learning with Applications in Breast Cancer Diagnosis and Prognosis." *Designs.*, 1–17. https://doi.org/10.3390/designs2020013.

Zhang, Baoshuang, Yanying Li, and Zheng Chai. 2022. "A Novel Random Multi-Subspace Based ReliefF for Feature Selection." *Knowledge-Based Systems* 252: 109400. https://doi.org/10.1016/j.knosys.2022.109400.

# Supplementary Material

# Towards improving Alzheimer's intervention: a machine learning approach for biomarker detection through combining MEG and MRI pipelines

Alwani Liyana Ahmad[1,2,3], Jose Sanchez-Bornot[4, *], Roberto C. Sotero[5], Damien Coyle[6], Zamzuri Idris[2,3,7], Ibrahima Faye[1,8, *]

[1] Department of Fundamental and Applied Sciences, Faculty of Science and Information Technology, Universiti Teknologi PETRONAS, Perak, Malaysia.
[2] Department of Neurosciences, Hospital Universiti Sains Malaysia, Kelantan, Malaysia.
[3] Brain and Behaviour Cluster, School of Medical Sciences, Universiti Sains Malaysia, Kelantan, Malaysia
[4] Intelligent Systems Research Centre, School of Computing, Engineering and Intelligent Systems, Ulster University, Magee campus, Derry~Londonderry, BT48 7JL, UK.
[5] Department of Radiology and Hotchkiss Brain Institute, University of Calgary, Calgary, AB, Canada.
[6] The Bath Institute for the Augmented Human, University of Bath, Bath, BA2 7AY, UK.
[7] Department of Neurosciences, School of Medical Sciences, Universiti Sains Malaysia, Kelantan, Malaysia
[8] Centre for Intelligent Signal & Imaging Research (CISIR), Universiti Teknologi PETRONAS, Perak, Malaysia.

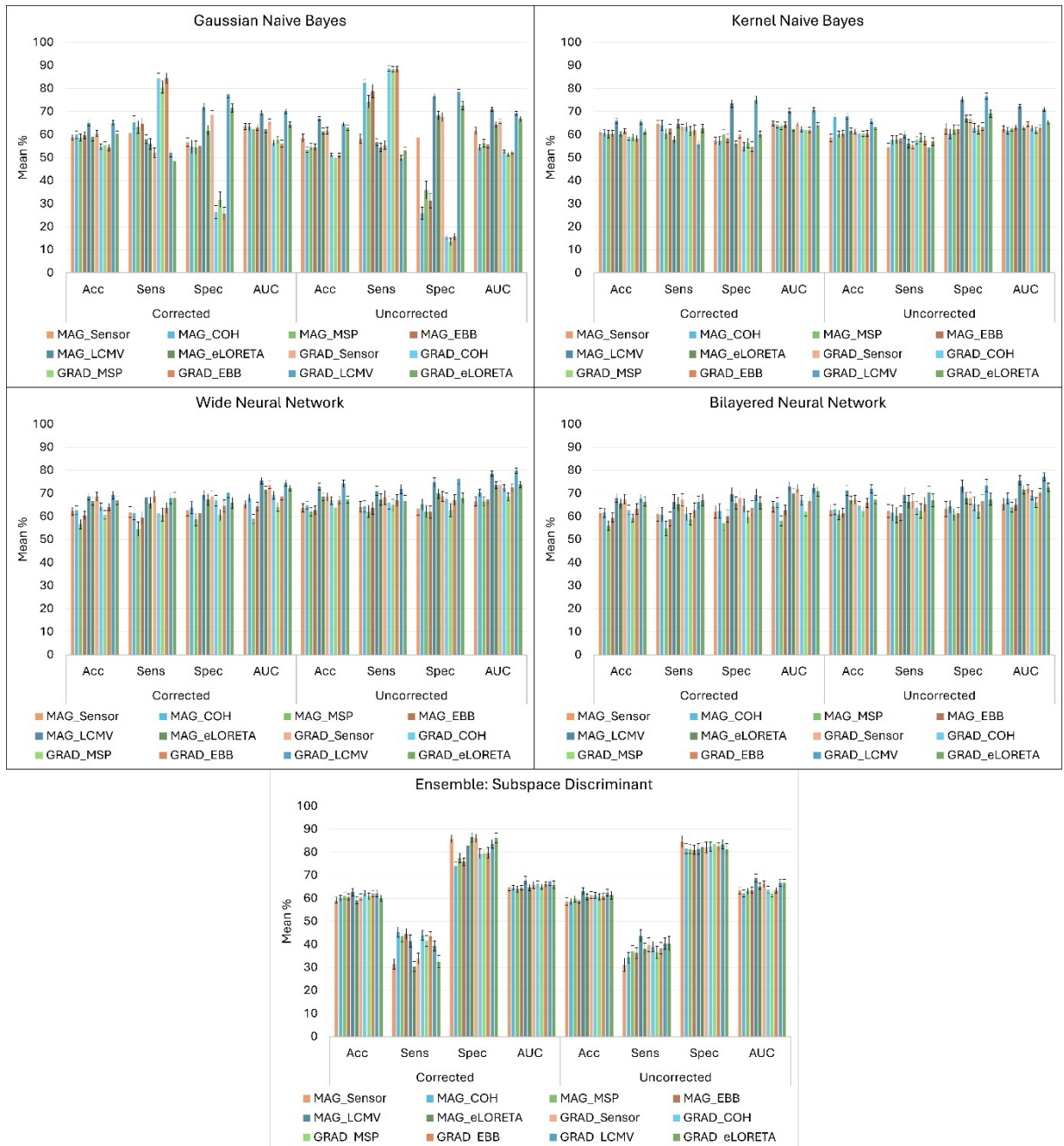

**Supplementary Figure 1:** Comparisons of five models; Gaussian Naïve Bayes, Kernel Naïve Bayes, Wide Neural Network, Bilayered Neural Network and Ensemble: Subspace Discriminant for corrected and uncorrected data in classification analysis using MEG-only based features.

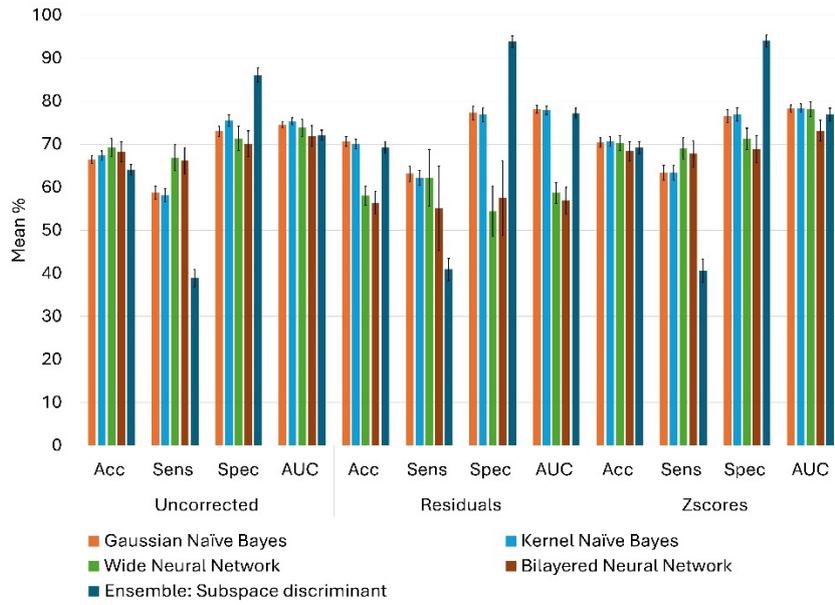

**Supplementary Figure 2**: Performance of MRI-only features across three different harmonization methods for Gaussian Naïve Bayes, Kernel Naïve Bayes, Wide Neural Network, Bilayered Neural Network and Ensemble: Subspace Discriminant

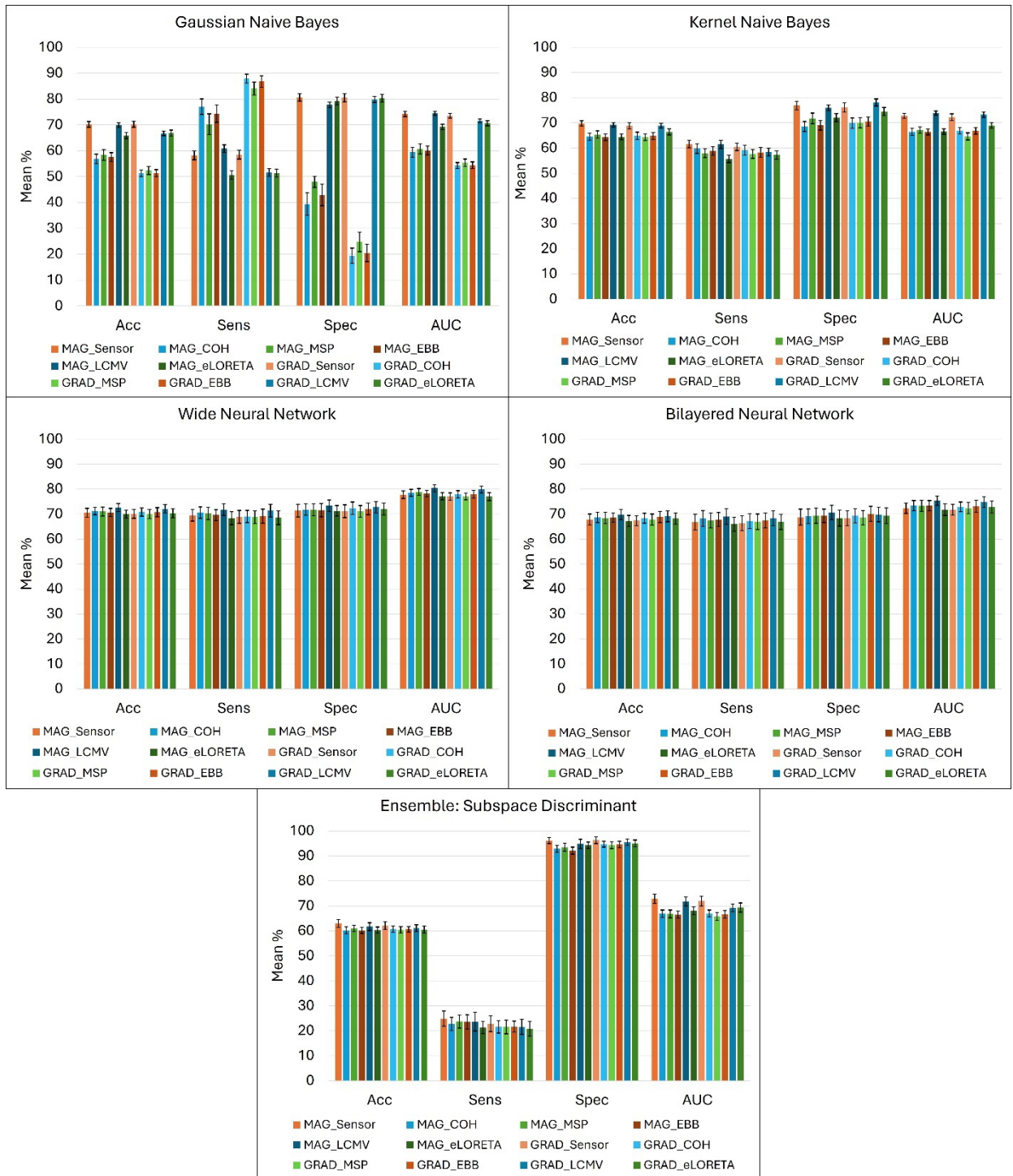

**Supplementary Figure 3**: Comparative performance of combined MAG MEG and MRI, and GRAD MEG and MRI using Gaussian Naïve Bayes, Kernel Naïve Bayes, Wide Neural Network, Bilayered Neural Network and Ensemble: Subspace Discriminant

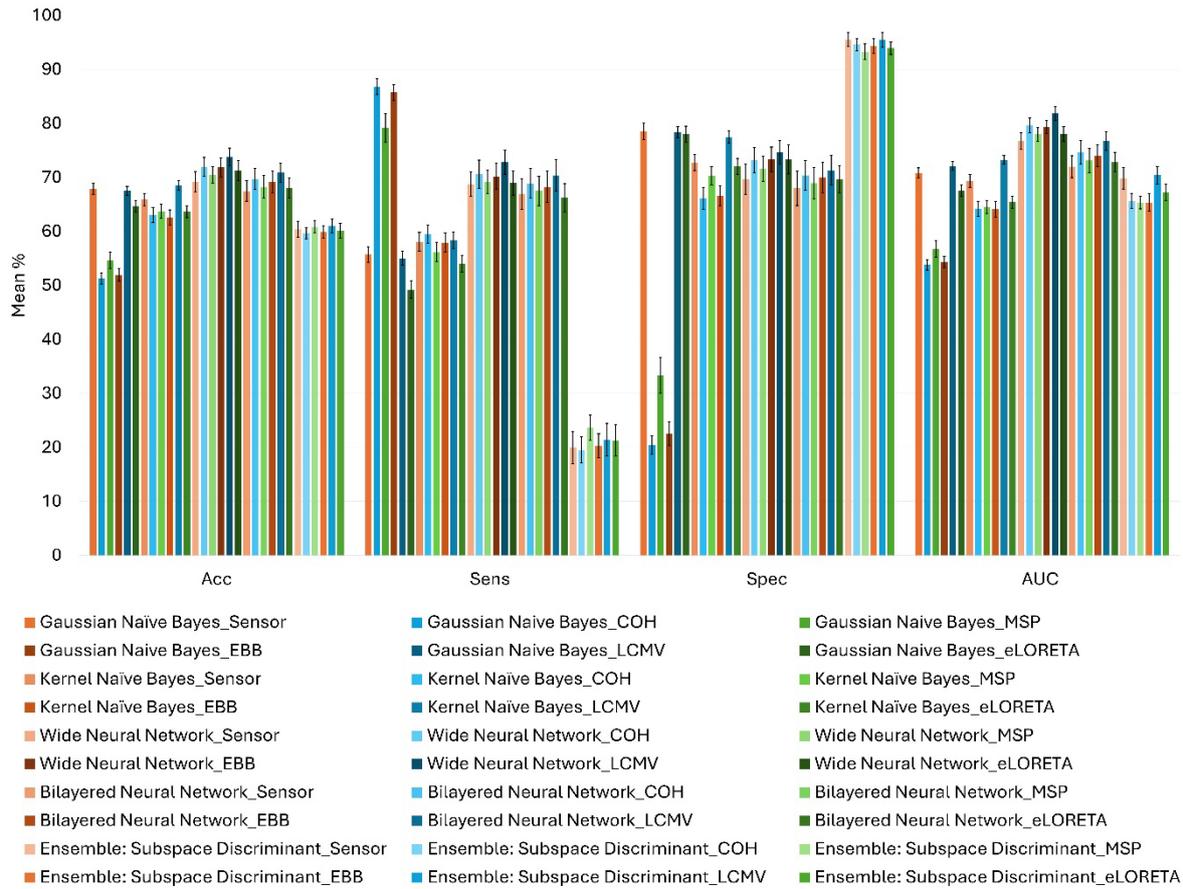

**Supplementary Figure 4:** Performance of combined MEG and MRI features using Gaussian Naïve Bayes, Kernel Naïve Bayes, Wide Neural Network, Bilayered Neural Network, and Ensemble: Subspace Discriminant

**Supplementary Table 1: Comparative statistics for the performance of MEG-only features.** Statistical summary of the N-way ANOVA assessing accuracy and AUC for the performance of MEG-only features across five different classifiers, influenced by both MEG sensors (MAG and GRAD), normalization methods, and source localization techniques.

***Classifier model and inverse solution methods pairwise comparison: post hoc test (Tukey's HSD)
*** Pairwise comparison: Adjustment for multiple comparison: Bonferroni
*** Gaussian Naïve Bayes: GNB, Kernel Naïve Bayes (KNB), Wide Neural Network (WNN), Bilayered Neural Network (BNN), & Ensemble: Subspace discriminant (ESD).

| Classification | classifier | modalities | Normalization | Source localization methods | Mean (SD) (%) | Contrast | p-value | 95% confidence interval | Pairwise comparison |
|---|---|---|---|---|---|---|---|---|---|
| Accuracy | Gaussian Naïve Bayes | MAGs | Corrected | Sensor | 58.7068 (1.2993) | GNB & KNB | $10^{-7}$ | [-0.0303, -0.0275] | GNB < KNB |
| | | | | COH | 59.8457 (1.7473) | GNB & WNN | $10^{-7}$ | [-0.0730, -0.0702] | GNB < WNN |
| | | | | MSP | 58.6574 (1.5104) | GNB & BNN | $10^{-7}$ | [-0.0609, -0.0581] | GNB < BNN |
| | | | | EBB | 59.6481 (1.4151) | GNB & ESD | $10^{-7}$ | [-0.0249, -0.0221] | GNB < ESD |
| | | | | LCMV | 65.0556 (1.2327) | GNB & KSVM | $10^{-7}$ | [-0.0742, -0.0714] | GNB < KSVM |
| | | | | eLORETA | 58.9198 (1.7263) | GNB & GLMNET | $10^{-7}$ | [-0.0622, -0.0594] | GNB < GLMNET |
| | | | Uncorrected | Sensor | 58.4568 (1.7629) | KNB & WNN | $10^{-7}$ | [-0.0441, -0.0414] | KNB < WNN |
| | | | | COH | 53.4290 (1.2164) | KNB & BNN | $10^{-7}$ | [-0.0321, -0.0293] | KNB < BNN |
| | | | | MSP | 54.5895 (1.5421) | KNB & ESD | $10^{-7}$ | [0.0040, 0.0068] | KNB > ESD |
| | | | | EBB | 54.4815 (1.3701) | KNB & KSVM | $10^{-7}$ | [-0.0454, -0.0426] | KNB < KSVM |
| | | | | LCMV | 66.9228 (0.8774) | KNB & GLMNET | $10^{-7}$ | [-0.0333, -0.0305] | KNB < GLMNET |
| | | | | eLORETA | 61.4753 (1.3352) | WNN & BNN | $10^{-7}$ | [0.010, 0.0135] | WNN > BN |
| | | GRDs | Corrected | Sensor | 60.4784 (1.3595) | WNN & ESD | $10^{-7}$ | [0.0468, 0.0496] | WNN > ESD |
| | | | | COH | 54.6883 (1.2249) | WNN & KSVM | $10^{-7}$ | [-0.0026, 0.0002] | WNN < KSVM |
| | | | | MSP | 55.5123 (1.4510) | WNN & GLMNET | $10^{-7}$ | [0.0094, 0.0122] | WNN > GLMNET |
| | | | | EBB | 54.3519 (1.3764) | BNN & ESD | $10^{-7}$ | [0.0347, 0.0375] | BNN > ESD |
| | | | | LCMV | 64.9938 (1.0594) | BNN & KSVM | $10^{-7}$ | [-0.0147, -0.0119] | BNN < KSVM |
| | | | | eLORETA | 60.2377 (1.2393) | ESD & KSVM | $10^{-7}$ | [-0.0508, -0.0480] | ESD < KSVM |
| | | | Uncorrected | Sensor | 61.6698 (1.4633) | ESD & GLMNET | $10^{-7}$ | [-0.0387, -0.0360] | ESD < GLMNET |
| | | | | COH | 51.2407 (0.7509) | KSVM & GLMNET | $10^{-7}$ | [0.0106, 0.0134] | KSVM > GLMNET |
| | | | | MSP | 49.9568 (0.8242) | Sensor & COH | $10^{-7}$ | [0.0129, 0.0154] | Sensor > COH |
| | | | | EBB | 51.1327 (0.7958) | Sensor & MSP | $10^{-7}$ | [0.0335, 0.0360] | Sensor > MSP |
| | | | | LCMV | 64.4691 (0.8080) | Sensor & EBB | $10^{-7}$ | [0.0181, 0.0206] | Sensor > EBB |
| | | | | eLORETA | 63.0062 (1.2157) | Sensor & LCMV | $10^{-7}$ | [-0.0531, -0.0506] | Sensor < LCMV |
| | Kernel Naïve Bayes | MAGs | Corrected | Sensor | 60.8488 (1.1551) | Sensor & eLORETA | $10^{-7}$ | [-0.0109, -0.0084] | Sensor < eLORETA |
| | | | | COH | 60.5247 (1.5482) | COH & MSP | $10^{-7}$ | [0.0193, 0.0218] | COH > MSP |
| | | | | MSP | 60.1512 (1.6308) | COH & EBB | $10^{-7}$ | [0.0039, 0.0064] | COH > EBB |
| | | | | EBB | 60.4136 (1.4056) | COH & LCMV | $10^{-7}$ | [-0.0673, -0.0648] | COH < LCMV |
| | | | | LCMV | 65.8364 (1.1403) | COH & eLORETA | $10^{-7}$ | [-0.0251, -0.0226] | COH < eLORETA |
| | | | | eLORETA | 60.1543 (1.1291) | MSP & EBB | $10^{-7}$ | [-0.0166, -0.0141] | MSP < EBB |
| | | | Uncorrected | Sensor | 58.5525 (1.5873) | MSP & LCMV | $10^{-7}$ | [-0.0878, -0.0853] | MSP < LCMV |
| | | | | COH | 59.0123 (1.5139) | MSP & eLORETA | $10^{-7}$ | [-0.0456, -0.0431] | MSP < eLORETA |
| | | | | MSP | 60.0494 (1.3805) | EBB & LCMV | $10^{-7}$ | [-0.0724, -0.0699] | EBB < LCMV |
| | | | | EBB | 60.3611 (1.3867) | EBB & eLORETA | $10^{-7}$ | [-0.0303, -0.0278] | EBB < eLORETA |
| | | | | LCMV | 67.8210 (1.1512) | LCMV & eLORETA | $10^{-7}$ | [0.0409, 0.0434] | LCMV > eLORETA |
| | | | | eLORETA | 61.5895 (1.2247) | ==MAG & GRAD== | ==$10^{-7}$== | ==$F_{(1, 16632)}$ = 1838.097== | ==MAG < GRAD== |
| | | GRDs | Corrected | Sensor | 61.5525 (1.0083) | Corrected & Uncorrected | $10^{-7}$ | $F_{(1, 16632)}$ = 1695.737 | Corrected < Uncorrected |
| | | | | COH | 58.9414 (1.3687) | | | | |
| | | | | MSP | 58.8025 (1.5237) | | | | |
| | | | | EBB | 58.1975 (1.5243) | | | | |
| | | | | LCMV | 65.5494 (1.0868) | | | | |
| | | | | eLORETA | 61.3951 (1.2288) | | | | |
| | | | Uncorrected | Sensor | 61.2469 (1.1484) | | | | |
| | | | | COH | 60.3519 (1.2950) | | | | |
| | | | | MSP | 60.3426 (1.1111) | | | | |
| | | | | EBB | 60.5031 (1.2778) | | | | |
| | | | | LCMV | 65.8148 (1.0429) | | | | |
| | | | | eLORETA | 63.1914 (1.1292) | | | | |
| | Wide Neural Network | MAGs | Corrected | Sensor | 62.3302 (1.7593) | | | | |
| | | | | COH | 62.6914 (1.8804) | | | | |
| | | | | MSP | 56.6914 (2.0817) | | | | |
| | | | | EBB | 60.4043 (1.7350) | | | | |
| | | | | LCMV | 68.7623 (1.5362) | | | | |
| | | | | eLORETA | 66.5617 (1.6838) | | | | |
| | | | Uncorrected | Sensor | 63.7037 (1.8135) | | | | |
| | | | | COH | 64.8179 (1.5783) | | | | |
| | | | | MSP | 62.1049 (1.8133) | | | | |
| | | | | EBB | 62.7500 (1.7562) | | | | |
| | | | | LCMV | 72.9568 (1.5614) | | | | |
| | | | | eLORETA | 68.5864 (1.6534) | | | | |
| | | GRDs | Corrected | Sensor | 68.6111 (1.8968) | | | | |
| | | | | COH | 64.1173 (1.7627) | | | | |
| | | | | MSP | 60.6512 (1.8464) | | | | |
| | | | | EBB | 64.1019 (1.5797) | | | | |
| | | | | LCMV | 69.2531 (1.7079) | | | | |
| | | | | eLORETA | 66.9630 (1.7509) | | | | |
| | | | Uncorrected | Sensor | 68.4660 (1.8104) | | | | |
| | | | | COH | 66.8704 (1.6889) | | | | |
| | | | | MSP | 63.8457 (1.7981) | | | | |
| | | | | EBB | 67.0617 (1.6178) | | | | |
| | | | | LCMV | 74.1512 (1.5574) | | | | |
| | | | | eLORETA | 67.3549 (1.5378) | | | | |
| | Bilayered Neural Network | MAGs | Corrected | Sensor | 61.4784 (2.1078) | | | | |
| | | | | COH | 61.6204 (2.0385) | | | | |

| | | | | MSP | 55.9691 (2.0285) | | | | |
|---|---|---|---|---|---|---|---|---|---|
| | | | | EBB | 59.5340 (1.9525) | | | | |
| | | | | LCMV | 68.0833 (1.9223) | | | | |
| | | | | eLORETA | 65.6420 (2.0298) | | | | |
| | | | Uncorrected | Sensor | 62.8025 (2.2763) | | | | |
| | | | | COH | 62.9815 (2.1436) | | | | |
| | | | | MSP | 60.6265 (2.0814) | | | | |
| | | | | EBB | 61.4475 (2.0749) | | | | |
| | | | | LCMV | 71.2500 (2.0451) | | | | |
| | | | | eLORETA | 67.1265 (1.7746) | | | | |
| | | GRDs | Corrected | Sensor | 67.5093 (2.0608) | | | | |
| | | | | COH | 62.8395 (2.0270) | | | | |
| | | | | MSP | 59.2500 (1.7240) | | | | |
| | | | | EBB | 63.2068 (2.3666) | | | | |
| | | | | LCMV | 67.8210 (1.8213) | | | | |
| | | | | eLORETA | 66.4414 (1.9588) | | | | |
| | | | Uncorrected | Sensor | 67.2191 (1.9423) | | | | |
| | | | | COH | 64.6049 (1.9176) | | | | |
| | | | | MSP | 62.2377 (2.0907) | | | | |
| | | | | EBB | 65.8827 (2.0119) | | | | |
| | | | | LCMV | 71.9660 (1.8482) | | | | |
| | | | | eLORETA | 67.2407 (1.8320) | | | | |
| | Ensemble: Subspace discriminant | MAGs | Corrected | Sensor | 59.2623 (1.2318) | | | | |
| | | | | COH | 60.0617 (1.2593) | | | | |
| | | | | MSP | 61.0000 (1.4893) | | | | |
| | | | | EBB | 60.6235 (1.3894) | | | | |
| | | | | LCMV | 62.6821 (1.6537) | | | | |
| | | | | eLORETA | 59.1667 (1.5111) | | | | |
| | | | Uncorrected | Sensor | 58.6235 (1.5877) | | | | |
| | | | | COH | 58.5185 (1.3284) | | | | |
| | | | | MSP | 59.6914 (1.4563) | | | | |
| | | | | EBB | 59.1852 (1.2632) | | | | |
| | | | | LCMV | 63.0370 (1.5818) | | | | |
| | | | | eLORETA | 60.6204 (1.3529) | | | | |

| Metric | Classifier | Sensor | Correction | Method | Value | Comparison | p | CI | Result |
|---|---|---|---|---|---|---|---|---|---|
| | | GRDs | Corrected | Sensor | 60.5772 (1.3520) | | | | |
| | | | | COH | 62.0432 (1.2969) | | | | |
| | | | | MSP | 60.9444 (1.3931) | | | | |
| | | | | EBB | 62.0432 (1.3168) | | | | |
| | | | | LCMV | 62.0278 (1.4473) | | | | |
| | | | | eLORETA | 60.0031 (1.4463) | | | | |
| | | | Uncorrected | Sensor | 61.4691 (1.5176) | | | | |
| | | | | COH | 61.3117 (1.3243) | | | | |
| | | | | MSP | 60.6481 (1.3652) | | | | |
| | | | | EBB | 60.9383 (1.2596) | | | | |
| | | | | LCMV | 62.3673 (1.4689) | | | | |
| | | | | eLORETA | 61.3735 (1.7517) | | | | |
| AUC | Gaussian Naïve Bayes | MAGs | Corrected | Sensor | 63.4846 (1.3008) | GNB & KNB | $10^{-7}$ | [-0.0317, -0.0289] | GNB < KNB |
| | | | | COH | 63.2008 (1.5194) | GNB & WNN | $10^{-7}$ | [-0.0908, -0.0881] | GNB < WNN |
| | | | | MSP | 62.4788 (1.4892) | GNB & BNN | $10^{-7}$ | [-0.0710, -0.0682] | GNB < BNN |
| | | | | EBB | 63.0499 (1.2074) | GNB & ESD | $10^{-7}$ | [-0.0378, -0.0350] | GNB < ESD |
| | | | | LCMV | 69.3886 (1.1222) | GNB & KSVM | $10^{-7}$ | [-0.0990, -0.0962] | GNB < KSVM |
| | | | | eLORETA | 62.2345 (1.3502) | GNB & GLMNET | $10^{-7}$ | [-0.0689, -0.0662] | GNB < GLMNET |
| | | | Uncorrected | Sensor | 61.5947 (1.6415) | KNB & WNN | $10^{-7}$ | [-0.0605, -0.0577] | KNB < WNN |
| | | | | COH | 54.3430 (1.2705) | KNB & BNN | $10^{-7}$ | [-0.0407, -0.0379] | KNB < BNN |
| | | | | MSP | 56.2304 (1.6359) | KNB & ESD | $10^{-7}$ | [-0.0075, -0.0047] | KNB < ESD |
| | | | | EBB | 55.6317 (1.3714) | KNB & KSVM | $10^{-7}$ | [-0.0686, -0.0659] | KNB < KSVM |
| | | | | LCMV | 70.9221 (0.8241) | KNB & eLORETA | $10^{-7}$ | [-0.0386, -0.0358] | KNB < eLORETA |
| | | | | eLORETA | 64.4382 (1.2090) | WNN & BNN | $10^{-7}$ | [0.0185, 0.0212] | WNN > BNN |
| | | GRDs | Corrected | Sensor | 65.5741 (1.2241) | WNN & ESD | $10^{-7}$ | [0.0517, 0.0544] | WNN > ESD |
| | | | | COH | 56.4147 (1.3336) | WNN & KSVM | $10^{-7}$ | [-0.0095, 0.0067] | WNN < KSVM |
| | | | | MSP | 57.6697 (1.6187) | WNN & eLORETA | $10^{-7}$ | [0.0205, 0.0233] | WNN > eLORETA |
| | | | | EBB | 56.0709 (1.4711) | BNN & ESD | $10^{-7}$ | [0.0318, 0.0346] | BNN > ESD |
| | | | | LCMV | 70.0864 (1.0812) | BNN & KSVM | $10^{-7}$ | [-0.0294, -0.0266] | BNN < KSVM |
| | | | | eLORETA | 64.2125 (1.1925) | BNN & eLORETA | < 0.001 | [0.0007, 0.0034] | BNN > eLORETA |
| | | | Uncorrected | Sensor | 65.7836 (1.0807) | ESD & KSVM | $10^{-7}$ | [-0.0626, -0.0598] | ESD < KSVM |
| | | | | COH | 52.4757 (0.7337) | ESD & GLMNET | $10^{-7}$ | [-0.0325, -0.0298] | ESD < GLMNET |
| | | | | MSP | 51.3114 (0.7958) | KSVM & GLMNET | $10^{-7}$ | [0.0286, 0.0314] | KSVM > GLMNET |
| | | | | EBB | 52.3106 (0.8121) | Sensor & COH | $10^{-7}$ | [0.0164, 0.0189] | Sensor > COH |

| | | | | Method | Value | Comparison | p | CI / F | Result |
|---|---|---|---|---|---|---|---|---|---|
| | | | | LCMV | 69.2996 (0.8014) | Sensor & MSP | $10^{-7}$ | [0.0448, 0.0473] | Sensor > MSP |
| | | | | eLORETA | 66.7483 (1.1943) | Sensor & EBB | $10^{-7}$ | [0.0238, 0.0263] | Sensor > EBB |
| | Kernel Naïve Bayes | MAGs | Corrected | Sensor | 64.8267 (0.9497) | Sensor & LCMV | $10^{-7}$ | 0.0640, -0.0616 | Sensor > LCMV |
| | | | | COH | 64.1118 (1.2707) | Sensor & eLORETA | $10^{-7}$ | [-0.0127, -0.0102] | Sensor < eLORETA |
| | | | | MSP | 63.5451 (1.4292) | COH & MSP | $10^{-7}$ | [0.0271, 0.0296] | COH > MSP |
| | | | | EBB | 64.2708 (1.1864) | COH & EBB | $10^{-7}$ | [0.0061, 0.0086] | COH > EBB |
| | | | | LCMV | 70.3601 (0.9715) | COH & LCMV | $10^{-7}$ | [-0.0817, -0.0792] | COH < LCMV |
| | | | | eLORETA | 62.4420 (0.8630) | COH & eLORETA | $10^{-7}$ | [-0.0304, -0.0279] | COH < eLORETA |
| | | | Uncorrected | Sensor | 62.3920 (1.4247) | MSP & EBB | $10^{-7}$ | [-0.0223, -0.0198] | MSP < EBB |
| | | | | COH | 61.6338 (1.4190) | MSP & LCMV | $10^{-7}$ | [-0.1101, -0.1076] | MSP < LCMV |
| | | | | MSP | 62.7403 (1.2358) | MSP & eLORATA | $10^{-7}$ | [-0.0588, -0.0563] | MSP < eLORATA |
| | | | | EBB | 62.9698 (1.3327) | EBB & LCMV | $10^{-7}$ | [-0.0891, -0.0866] | EBB < LCMV |
| | | | | LCMV | 72.0925 (0.8847) | EBB & eLORETA | $10^{-7}$ | [-0.0377, -0.0352] | EBB < eLORETA |
| | | | | eLORETA | 63.0051 (0.9573) | LCMV & eLORETA | $10^{-7}$ | [0.0501, 0.0526] | LCMV > eLORETA |
| | | GRDs | Corrected | Sensor | 64.4586 (0.8192) | MAG & GRAD | $10^{-7}$ | $F_{(1,\ 16632)} = 2420.650$ | MAG < GRAD |
| | | | | COH | 62.3745 (1.2278) | Corrected & Uncorrected | < 0.001 | $F_{(1,\ 16632)} = 1329.410$ | Corrected < Uncorrected |
| | | | | MSP | 61.9938 (1.3666) | | | | |
| | | | | EBB | 61.9690 (1.2089) | | | | |
| | | | | LCMV | 70.6350 (1.0696) | | | | |
| | | | | eLORETA | 64.0295 (1.0505) | | | | |
| | | | Uncorrected | Sensor | 64.4656 (1.0376) | | | | |
| | | | | COH | 62.6164 (1.3243) | | | | |
| | | | | MSP | 61.7542 (1.3142) | | | | |
| | | | | EBB | 62.8295 (1.1781) | | | | |
| | | | | LCMV | 70.9411 (0.9384) | | | | |
| | | | | eLORETA | 65.2907 (0.9332) | | | | |
| | Wide Neural Network | MAGs | Corrected | Sensor | 65.1901 (1.8201) | | | | |
| | | | | COH | 67.9675 (1.7102) | | | | |
| | | | | MSP | 58.9515 (1.9423) | | | | |
| | | | | EBB | 64.2818 (1.8267) | | | | |
| | | | | LCMV | 75.5072 (1.3778) | | | | |
| | | | | eLORETA | 71.6634 (1.4358) | | | | |
| | | | Uncorrected | Sensor | 66.5213 (2.0223) | | | | |
| | | | | COH | 70.2475 (1.5616) | | | | |

| | | | | | | | | |
|---|---|---|---|---|---|---|---|---|
| | | | | MSP | 66.5126 (1.8885) | | | |
| | | | | EBB | 67.2928 (1.6550) | | | |
| | | | | LCMV | 78.5693 (1.1775) | | | |
| | | | | eLORETA | 73.6120 (1.4739) | | | |
| | | GRDs | Corrected | Sensor | 73.7193 (1.5955) | | | |
| | | | | COH | 69.2387 (1.6998) | | | |
| | | | | MSP | 64.1727 (1.8769) | | | |
| | | | | EBB | 68.7148 (1.5758) | | | |
| | | | | LCMV | 74.4782 (1.4319) | | | |
| | | | | eLORETA | 72.3394 (1.3163) | | | |
| | | | Uncorrected | Sensor | 73.6340 (1.6658) | | | |
| | | | | COH | 72.1469 (1.5578) | | | |
| | | | | MSP | 68.5298 (1.6291) | | | |
| | | | | EBB | 72.6823 (1.4868) | | | |
| | | | | LCMV | 79.8706 (1.1739) | | | |
| | | | | eLORETA | 73.7869 (1.3816) | | | |
| | Bilayered Neural Network | MAGs | Corrected | Sensor | 64.1973 (2.3496) | | | |
| | | | | COH | 66.0804 (2.1229) | | | |
| | | | | MSP | 58.1026 (2.2000) | | | |
| | | | | EBB | 62.7783 (2.0992) | | | |
| | | | | LCMV | 72.9164 (1.9442) | | | |
| | | | | eLORETA | 70.0867 (2.0952) | | | |
| | | | Uncorrected | Sensor | 65.4428 (2.4865) | | | |
| | | | | COH | 67.7882 (2.3016) | | | |
| | | | | MSP | 64.0491 (2.1184) | | | |
| | | | | EBB | 65.1284 (2.2483) | | | |
| | | | | LCMV | 75.7139 (2.0024) | | | |
| | | | | eLORETA | 71.7214 (1.8359) | | | |
| | | GRDs | Corrected | Sensor | 72.0714 (1.7860) | | | |
| | | | | COH | 67.0060 (2.0817) | | | |
| | | | | MSP | 62.1807 (1.8829) | | | |
| | | | | EBB | 66.6266 (2.3377) | | | |
| | | | | LCMV | 72.2481 (2.0229) | | | |
| | | | | eLORETA | 70.8745 (2.0759) | | | |
| | | | Uncorrected | Sensor | 71.8765 (1.9161) | | | |
| | | | | COH | 69.0844 (1.9453) | | | |
| | | | | MSP | 66.0588 (2.1581) | | | |
| | | | | EBB | 70.2286 (1.8968) | | | |
| | | | | LCMV | 77.1263 (1.7862) | | | |
| | | | | eLORETA | 72.6295 (1.7604) | | | |
| | Ensemble: Subspace Discriminant | MAGs | Corrected | Sensor | 64.8324 (1.5490) | | | |
| | | | | COH | 64.5262 (1.0270) | | | |
| | | | | MSP | 64.0339 (1.3235) | | | |
| | | | | EBB | 64.5715 (0.9427) | | | |
| | | | | LCMV | 67.7758 (1.6690) | | | |
| | | | | eLORETA | 64.6622 (1.4002) | | | |
| | | | Uncorrected | Sensor | 63.4888 (1.5158) | | | |
| | | | | COH | 62.0829 (1.5193) | | | |
| | | | | MSP | 63.0932 (1.3384) | | | |
| | | | | EBB | 63.5349 (1.3250) | | | |
| | | | | LCMV | 68.8510 (1.6520) | | | |
| | | | | eLORETA | 65.2223 (1.2763) | | | |
| | | GRDs | Corrected | Sensor | 65.8126 (1.5417) | | | |
| | | | | COH | 66.2704 (1.2233) | | | |
| | | | | MSP | 64.9929 (1.1022) | | | |
| | | | | EBB | 66.1754 (1.1636) | | | |
| | | | | LCMV | 67.5018 (1.8765) | | | |
| | | | | eLORETA | 65.8284 (1.6604) | | | |
| | | | Uncorrected | Sensor | 66.2021 (1.4448) | | | |
| | | | | COH | 63.7827 (1.4070) | | | |
| | | | | MSP | 62.2220 (1.5325) | | | |
| | | | | EBB | 63.4778 (1.2137) | | | |
| | | | | LCMV | 66.7528 (1.5171) | | | |
| | | | | eLORETA | 66.6218 (1.5903) | | | |

**Supplementary Table 2: Comparative statistics for the performance of MRI-only features.** Statistical summary of the N-way ANOVA assessing accuracy and AUC for the performance of MRI-only features across seven different classifiers, influenced by normalization methods.

***Classifier model and inverse solution methods pairwise comparison: post hoc test (Tukey's HSD)

\*\*\* Pairwise comparison: Adjustment for multiple comparison: Bonferroni
\*\*\* Gaussian Naïve Bayes: GNB, Kernel Naïve Bayes (KNB), Wide Neural Network (WNN), Bilayered Neural Network (BNN), Ensemble: Subspace discriminant (ESD), & Kernel Support Vector Machine (KSVM)

| Classification | classifier | Normalization | Mean (SD) (%) | Contrast | p-value | 95% confidence interval | Pairwise comparison |
|---|---|---|---|---|---|---|---|
| Accuracy | Gaussian Naïve Bayes | Uncorrected | 66.4257 (0.9368) | GNB & WNN | < 0.001 | [0.0290, 0.0372] | GNB > WNN |
| | | Residuals | 70.6997 (1.1274) | GNB & BNN | < 0.001 | [0.0441, 0.0523] | GNB > BNN |
| | | Z-scores | 70.3993 (1.1323) | GNB & ESD | < 0.001 | [0.0126, 0.0208] | GNB > ESD |
| | Kernel Naïve Bayes | Uncorrected | 67.4191 (1.0331) | GNB & KSVM | < 0.001 | [0.0271, 0.0353] | GNB > KSVM |
| | | Residuals | 70.0495 (1.1398) | GNB & GLMNET | < 0.001 | [-0.0362, -0.0279] | GNB < GLMNET |
| | | Z-scores | 70.6238 (1.1409) | KNB & WNN | < 0.001 | [0.0309, 0.0391] | KNB > WNN |
| | Wide Neural Network | Uncorrected | 69.2640 (2.0908) | KNB & BNN | < 0.001 | [0.0460, 0.0542] | KNB > BNN |
| | | Residuals | 58.0759 (2.2273) | KNB & ESD | < 0.001 | [0.0145, 0.0227] | KNB > ESD |
| | | Z-scores | 70.2541 (1.7600) | KNB & KSVM | < 0.001 | [0.0290, 0.0372] | KNB > KSVM |
| | Bilayered Neural Network | Uncorrected | 68.2607 (2.2560) | KNB & GLMNET | < 0.001 | [-0.0343, -0.0261] | KNB < GLMNET |
| | | Residuals | 56.4026 (2.6645) | WNN & BNN | < 0.001 | [0.0110, 0.0192] | WNN > BNN |
| | | Z-scores | 68.3927 (2.2364) | WNN & ESD | < 0.001 | [-0.0205, -0.0123] | WNN < ESD |
| | Ensemble: Subspace discriminant | Uncorrected | 64.1221 (1.1550) | WNN & GLMNET | < 0.001 | [-0.0693, 0.0610] | WNN < GLMNET |
| | | Residuals | 69.2079 (1.3363) | BNN & ESD | < 0.001 | [-0.0356, -0.0274] | BNN < ESD |
| | | Z-scores | 69.1848 (1.3282) | BNN & KSVM | < 0.001 | [-0.0211, -0.0129] | BNN < KSVM |
| | Kernel Support Vector Machine | Uncorrected | 71.3432 (1.7217) | BNN & GLMNET | < 0.001 | [-0.0844, -0.0762] | BNN < GLMNET |
| | | Residuals | 55.0627 (2.8238) | ESD & KSVM | < 0.001 | [0.0104, 0.0186] | ESD > KSVM |
| | | Z-scores | 71.7525 (1.5323) | ESD & GLMNET | < 0.001 | [-0.0529, -0.0446] | ESD < GLMNET |
| | GLMNET | Uncorrected | 71.8218 (1.4585) | KSVM & GLMNET | < 0.001 | [-0.0674, -0.0592] | KSVM < GLMNET |
| | | Residuals | 72.5842 (1.5475) | Uncorrected & Residuals | < 0.001 | [0.0358, 0.0401] | Uncorrected > Residuals |
| | | Z-scores | 72.7360 (1.3350) | Uncorrected & Z-scores | < 0.001 | [-0.0231, -0.0188] | Uncorrected < Z-scores |
| | | | | Residuals & Z-scores | < 0.001 | [-0.0611, -0.0568] | Residuals < Z-scores |
| AUC | Gaussian Naïve Bayes | Uncorrected | 74.5627 (0.6900) | GNB & WNN | < 0.001 | [0.0632, 0.0720] | GNB > WNN |
| | | Residuals | 78.1607 (0.9265) | GNB & BNN | < 0.001 | [0.0924, 0.1011] | GNB > BNN |
| | | Z-scores | 78.2573 (0.8782) | GNB & ESD | < 0.001 | [0.0113, 0.0200] | GNB > ESD |
| | Kernel Naïve Bayes | Uncorrected | 75.3337 (0.8004) | GNB & KSVM | < 0.001 | [0.0610, 0.0697] | GNB > KSVM |
| | | Residuals | 77.8979 (0.9595) | GNB & GLMNET | < 0.001 | [-0.0278, -0.0190] | GNB < GLMNET |
| | | Z-scores | 78.4204 (1.0597) | KNB & WNN | < 0.001 | [0.0655, 0.0742] | KNB > WNN |
| | Wide Neural Network | Uncorrected | 73.8520 (2.0039) | KNB & BNN | < 0.001 | [0.0947, 0.1034] | KNB > WNN |
| | | Residuals | 58.7114 (2.4516) | KNB & ESD | < 0.001 | [0.0136, 0.0223] | KNB > ESD |
| | | Z-scores | 78.1377 (1.6507) | KNB & KSVM | < 0.001 | [0.0632, 0.0719] | KNB > KSVM |
| | Bilayered Neural Network | Uncorrected | 71.9497 (2.3581) | KNB & GLMNET | < 0.001 | [-0.0255, -0.0168] | KNB < GLMNET |
| | | Residuals | 56.8838 (3.1050) | WNN & BNN | < 0.001 | [0.0248, 0.0336] | WNN > BNN |
| | | Z-scores | 73.1120 (2.4276) | WNN & ESD | < 0.001 | [-0.0563, -0.0476] | WNN < ESD |
| | Ensemble: Subspace discriminant | Uncorrected | 72.1214 (1.1761) | WNN & GLMNET | < 0.001 | [-0.0953, -0.0866] | WNN < GLMNET |
| | | Residuals | 77.2228 (1.1381) | BNN & ESD | < 0.001 | [-0.0855, -0.0767] | BNN < ESD |
| | | Z-scores | 76.9315 (1.4136) | BNN & KSVM | < 0.001 | [-0.0358, -0.0271] | BNN < KSVM |
| | Kernel Support Vector Machine | Uncorrected | 78.7920 (1.6796) | BNN & GLMNET | < 0.001 | [-0.1245, -0.1158] | BNN < GLMNET |
| | | Residuals | 54.0024 (4.0032) | ESD & KSVM | < 0.001 | [0.0453, 0.0540] | ESD > KSVM |
| | | Z-scores | 78.5896 (1.4538) | ESD & GLMNET | < 0.001 | [-0.0434, -0.0347] | ESD < GLMNET |
| | GLMNET | Uncorrected | 79.4634 (0.9868) | KSVM & GLMNET | < 0.001 | [-0.0931, -0.0843] | KSVM < GLMNET |
| | | Residuals | 79.2125 (1.3568) | Uncorrected & Residuals, | < 0.001 | [0.0606, 0.0651] | Uncorrected > Residuals |

| | | Z-scores | 79.3206 (1.3501) | Uncorrected & Z-scores | < 0.001 | [-0.0261, -0.0216] | Uncorrected < Z-scores |
| | | | | Residuals & Z-scores | < 0.001 | [-0.0890, -0.0844] | Residuals < Z-scores |

**Supplementary Table 3: Comparative statistics for combining uncorrected MEG MAG with MRI z-scores and uncorrected MEG GRAD with MRI z-scores.** Statistical summary of the N-way ANOVA assessing accuracy and AUC for the performance of combination MEG MAG with MRI and MEG GRAD with MRI features across five different classifiers, influenced by both MEG sensors (MAG and GRAD) and source localization techniques.

***Classifier model and inverse solution methods pairwise comparison: p value post hoc test (Tukey's HSD)
*** Pairwise comparison: p-value: Adjustment for multiple comparison: Bonferroni
*** Gaussian Naïve Bayes: GNB, Kernel Naïve Bayes (KNB), Wide Neural Network (WNN), Bilayered Neural Network (BNN), & Ensemble: Subspace discriminant (ESD)

| Classification | Classifier | Modalities | Source localization methods | Mean (SD) (%) | Contrast | p-value | 95% confidence interval | Pairwise comparison |
|---|---|---|---|---|---|---|---|---|
| Accuracy | Gaussian Naïve Bayes | MAG | Sensor | 70.2178 (1.1092) | GNB & KNB | < 0.001 | [-0.0520, -0.0447] | GNB < KNB |
| | | | COH | 56.9043 (1.8017) | GNB & WNN | < 0.001 | [-0.0973, -0.0901] | GNB < WNN |
| | | | MSP | 58.3597 (2.0731) | GNB & BNN | < 0.001 | [-0.0725, -0.0653] | GNB < BNN |
| | | | EBB | 57.5512 (1.7028) | GNB & ESD | 0.003 | [0.0011, 0.0083] | GNB > ESD |
| | | | LCMV | 69.8713 (0.9112) | GNB & KSVM | < 0.001 | [-0.1110, -0.1037] | GNB < KSVM |
| | | | eLORETA | 65.9274 (1.1482) | GNB & GLMNET | < 0.001 | [-0.1275, -0.1203] | GNB < GLMNET |
| | | GRAD | Sensor | 70.2541 (1.1554) | KNN & WNN | < 0.001 | [-0.0490, -0.0417] | KNN < WNN |
| | | | COH | 51.3102 (1.2635) | KNN & BNN | < 0.001 | [-0.0241, -0.0169] | KNN < BNN |
| | | | MSP | 52.3663 (1.5280) | KNN & ESD | < 0.001 | [0.0495, 0.0567] | KNN > ESD |
| | | | EBB | 51.3597 (1.3618) | KNN & KSVM | < 0.001 | [-0.0626, -0.0553] | KNN < KSVM |
| | | | LCMV | 66.7228 (0.8650) | KNN & GLMNET | < 0.001 | [-0.0792, -0.0719] | KNN < GLMNET |
| | | | eLORETA | 66.8416 (1.1370) | WNN & BNN | < 0.001 | [0.0212, 0.0285] | WNN > BNN |
| | Kernel Naïve Bayes | MAG | Sensor | 69.7789 (1.0534) | WNN & ESD | < 0.001 | [0.0948, 0.1021] | WNN > ESD |
| | | | COH | 64.4818 (1.4296) | WNN & KSVM | < 0.001 | [-0.0173, -0.0100] | WNN < KSVM |
| | | | MSP | 65.3300 (1.4844) | WNN & GLMNET | < 0.001 | [-0.0338, -0.0266] | WNN < GLMNET |
| | | | EBB | 64.3168 (1.3036) | BNN & ESD | < 0.001 | [0.0700, 0.0772] | BNN > ESD |
| | | | LCMV | 69.2178 (0.8886) | BNN & KSVM | < 0.001 | [-0.0421, -0.0348] | BNN < KSVM |
| | | | eLORETA | 64.4158 (1.1223) | BNN & GLMNET | < 0.001 | [-0.0586, -0.0514] | BNN < GLMNET |
| | | GRAD | Sensor | 68.8350 (1.2650) | ESD & KSVM | < 0.001 | [-0.1157, -0.1084] | ESD < KSVM |
| | | | COH | 64.8977 (1.4447) | ESD & GLMNET | < 0.001 | [-0.1322, -0.1250] | ESD < GLMNET |
| | | | MSP | 64.2904 (1.2761) | KSVM & GLMNET | < 0.001 | [-0.0202, -0.0129] | KSVM < GLMNET |
| | | | EBB | 64.8185 (1.3700) | Sensor & COH | $10^{-7}$ | [0.0266, 0.0331] | Sensor > COH |
| | | | LCMV | 68.9241 (0.9633) | Sensor & MSP | $10^{-7}$ | [0.0269, 0.0334] | Sensor > MSP |
| | | | eLORETA | 66.4356 (1.2037) | Sensor & EBB | $10^{-7}$ | [0.0266, 0.0331] | Sensor > EBB |

| | Model | Type | Method | Mean (SD) | Comparison | p | CI | Result |
|---|---|---|---|---|---|---|---|---|
| | Wide Neural Network | MAG | Sensor | 70.5182 (1.7558) | Sensor & LCMV | 0.047 | [-0.0065, 2.9 x 10$^{-5}$] | Sensor < LCMV |
| | | | COH | 71.2310 (1.5215) | Sensor & eLORETA | 10$^{-7}$ | [0.0091, 0.0156] | Sensor > eLORETA |
| | | | MSP | 71.0132 (1.7621) | COH & LCMV | 10$^{-7}$ | [-0.0363, -0.0298] | COH < LCMV |
| | | | EBB | 70.6568 (1.6486) | COH & eLORETA | 10$^{-7}$ | [-0.0207, -0.0142] | COH < eLORETA |
| | | | LCMV | 72.6073 (1.5817) | MSP & LCMV | 10$^{-7}$ | [-0.0367, -0.0302] | MSP < LCMV |
| | | | eLORETA | 69.9673 (1.6101) | MSP & eLORETA | 10$^{-7}$ | [-0.0210, -0.0145] | MSP < eLORETA |
| | | GRAD | Sensor | 70.0957 (1.7966) | EBB & LCMV | 10$^{-7}$ | [-0.0364, -0.0299] | EBB < LCMV |
| | | | COH | 70.8317 (1.7106) | EBB & eLORETA | 10$^{-7}$ | [-0.0207, -0.0142] | EBB < eLORETA |
| | | | MSP | 70.0000 (1.8110) | LCMV & eLORETA | 10$^{-7}$ | [0.0124, 0.0189] | LCMV > eLORETA |
| | | | EBB | 70.7426 (1.8619) | MAG & GRAD | < 0.001 | $F_{(1, 8387)} = 109.902$ | MAG > GRAD |
| | | | LCMV | 72.1155 (1.6727) | | | | |
| | | | eLORETA | 70.3861 (1.8509) | | | | |
| | Bilayered Neural Network | MAG | Sensor | 67.8284 (2.2194) | | | | |
| | | | COH | 68.6898 (2.1149) | | | | |
| | | | MSP | 68.4389 (2.3074) | | | | |
| | | | EBB | 68.6073 (1.8344) | | | | |
| | | | LCMV | 69.8812 (2.0622) | | | | |
| | | | eLORETA | 67.2574 (2.1999) | | | | |
| | | GRAD | Sensor | 67.4026 (1.9718) | | | | |
| | | | COH | 68.3597 (1.9717) | | | | |
| | | | MSP | 67.7954 (2.2108) | | | | |
| | | | EBB | 68.8449 (2.2389) | | | | |
| | | | LCMV | 69.0759 (2.1794) | | | | |
| | | | eLORETA | 68.1716 (2.2415) | | | | |
| | Ensemble: Subspace discriminant | MAG | Sensor | 63.0165 (1.5228) | | | | |
| | | | COH | 60.2409 (1.4131) | | | | |
| | | | MSP | 61.0000 (1.2890) | | | | |
| | | | EBB | 60.2310 (1.2574) | | | | |
| | | | LCMV | 61.7228 (1.5586) | | | | |
| | | | eLORETA | 60.3300 (1.2429) | | | | |
| | | GRAD | Sensor | 62.1089 (1.5612) | | | | |
| | | | COH | 60.6733 (1.2541) | | | | |
| | | | MSP | 60.4488 (1.3166) | | | | |
| | | | EBB | 60.6601 (1.1071) | | | | |
| | | | LCMV | 61.0990 (1.3714) | | | | |
| | | | eLORETA | 60.5050 (1.3900) | | | | |
| AUC | Gaussian Naïve Bayes | MAG | Sensor | 74.2183 (1.0074) | GNB & KNB | $10^{-7}$ | [-0.0419, -0.0380] | GNB < KNB |
| | | | COH | 59.4475 (1.8837) | GNB & WNN | $10^{-7}$ | = [-0.1350, -0.1312] | GNB < WNN |
| | | | MSP | 60.7381 (1.9449) | GNB & BNN | $10^{-7}$ | [-0.0846, -0.0808] | GNB < BNN |
| | | | EBB | 60.0834 (1.6778) | GNB & ESD | $10^{-7}$ | [-0.0392, -0.0354] | GNB < ESD |
| | | | LCMV | 74.4891 (0.8052) | GNB & KSVM | $10^{-7}$ | [-0.1469, -0.1430] | GNB < KSVM |
| | | | eLORETA | 69.2596 (1.0137) | GNB & GLMNET | $10^{-7}$ | [-0.1521, -0.1483] | GNB < GLMNET |
| | | GRAD | Sensor | 73.5494 (0.9533) | KNB & WNN | $10^{-7}$ | [-0.0951, -0.0912] | KNB < BNN |
| | | | COH | 54.2949 (1.1453) | KNB & BNN | $10^{-7}$ | [-0.04467, -0.0408] | KNB > ESD |
| | | | MSP | 55.3494 (1.4603) | KNB & ESD | < 0.001 | [0.0007, 0.0046] | KNB > ESD |
| | | | EBB | 54.4003 (1.3440) | KNB & KSVM | $10^{-7}$ | [-0.1069, -0.1031] | KNB < KSVM |
| | | | LCMV | 71.5884 (0.7671) | KNB & GLMNET | $10^{-7}$ | [-0.1121, -0.1083] | KNB < GLMNET |
| | | | eLORETA | 70.6196 (0.9812) | WNN & BNN | $10^{-7}$ | [0.0485, 0.0523] | WNN > BNN |
| | Kernel Naïve Bayes | MAG | Sensor | 72.8068 (1.0000) | WNN & ESD (p = $10^{-7}$, 95% C.I = | $10^{-7}$ | [0.0939, 0.0977] | WNN > ESD |
| | | | COH | 66.4319 (1.3409) | WNN & KSVM (p = $10^{-7}$, 95% C.I = | $10^{-7}$ | [-0.0138, -0.0099] | WNN < KSVM |
| | | | MSP | 67.1635 (1.2291) | WNN & GLMNET (p = $10^{-7}$, 95% C.I = | $10^{-7}$ | [-0.0190, -0.0152] | WNN < GLMNET |
| | | | EBB | 66.4043 (1.1752) | BNN & ESD (p = $10^{-7}$, 95% C.I = | $10^{-7}$ | [0.0435, 0.0473] | BNN > ESD |
| | | | LCMV | 73.8619 (0.8316) | BNN & KSVM (p = $10^{-7}$, 95% C.I = | $10^{-7}$ | [-0.0642, -0.0603] | BNN < KSVM |
| | | | eLORETA | 66.5172 (1.1117) | BNN & GLMNET | $10^{-7}$ | [-0.0694, -0.0656] | BNN < GLMNET |
| | | GRAD | Sensor | 72.3062 (1.2977) | ESD & KSVM | $10^{-7}$ | [-0.1096, -0.1058] | ESD < KSVM |
| | | | COH | 66.9115 (1.2389) | ESD & GLMNET | $10^{-7}$ | [-0.1148, -0.1110] | ESD < GLMNET |
| | | | MSP | 64.5932 (1.3912) | KSVM & GLMNET | $10^{-7}$ | [-0.0071, -0.0033] | KSVM < GLMNET |
| | | | EBB | 66.7760 (1.2965) | Sensor & COH | $10^{-7}$ | [0.0351, 0.0385] | Sensor > COH |
| | | | LCMV | 73.2504 (1.0876) | Sensor & MSP | $10^{-7}$ | [0.0372, 0.0406] | Sensor > MSP |
| | | | eLORETA | 68.9584 (1.1322) | Sensor & EBB | $10^{-7}$ | [0.0353, 0.0388] | Sensor > EBB |
| | Wide Neural Network | MAG | Sensor | 77.8047 (1.5330) | Sensor & LCMV | $10^{-7}$ | [-0.0095, -0.0061] | Sensor < LCMV |
| | | | COH | 78.5310 (1.4030) | Sensor & eLORETA | $10^{-7}$ | [0.0141, 0.0175] | Sensor > eLORETA |
| | | | MSP | 78.8902 (1.3203) | COH & MSP | 0.006 | [0.0004, 0.0038] | COH > MSP |
| | | | EBB | 78.2523 (1.2785) | COH & LCMV | $10^{-7}$ | [-0.0463, -0.0429] | COH < LCMV |
| | | | LCMV | 80.3676 (1.4066) | COH & eLORETA | $10^{-7}$ | [-0.02269, -0.0193] | COH < eLORETA |
| | | | eLORETA | 77.1438 (1.4155) | MSP & EBB | 0.026 | [-0.0035, -0.0001] | MSP < EBB |
| | | GRAD | Sensor | 77.0156 (1.4904) | MSP & LCMV | $10^{-7}$ | [-0.0484, -0.0450] | MSP < LCMV |
| | | | COH | 77.9141 (1.4633) | MSP & eLORETA | $10^{-7}$ | [-0.0248, -0.0214] | MSP < eLORETA |
| | | | MSP | 77.0589 (1.3753) | EBB & LCMV | $10^{-7}$ | [-0.0466, -0.0431] | EBB < LCMV |
| | | | EBB | 77.9157 (1.5324) | EBB & eLORETA | $10^{-7}$ | [-0.0230, -0.0195] | EBB < eLORETA |
| | | | LCMV | 79.8489 (1.3429) | LCMV & eLORETA | $10^{-7}$ | [0.0219, 0.0253] | LCMV > eLORETA |
| | | | eLORETA | 77.0200 (1.5749) | MAG & GRAD | < 0.001 | $F_{(1, 8316)} = 509.647$ | MAG > GRAD |
| | Bilayered Neural Network | MAG | Sensor | 72.3808 (2.0843) | | | | |
| | | | COH | 73.3753 (2.0798) | | | | |
| | | | MSP | 73.2965 (2.2075) | | | | |
| | | | EBB | 73.3566 (2.0388) | | | | |
| | | | LCMV | 75.4080 (1.9045) | | | | |
| | | | eLORETA | 71.7615 (2.3015) | | | | |
| | | GRAD | Sensor | 71.7322 (2.1576) | | | | |
| | | | COH | 72.8783 (1.9765) | | | | |
| | | | MSP | 72.3942 (2.2679) | | | | |
| | | | EBB | 73.1377 (2.3500) | | | | |
| | | | LCMV | 74.7866 (2.1324) | | | | |
| | | | eLORETA | 72.7839 (2.3851) | | | | |
| | Ensemble: Subspace discriminant | MAG | Sensor | 72.8455 (1.8954) | | | | |
| | | | COH | 66.8531 (1.4757) | | | | |
| | | | MSP | 66.7793 (1.5472) | | | | |
| | | | EBB | 66.5345 (1.4207) | | | | |
| | | | LCMV | 71.7893 (1.7931) | | | | |
| | | | eLORETA | 68.0998 (1.5429) | | | | |
| | | GRAD | Sensor | 71.9330 (1.8989) | | | | |
| | | | COH | 66.9450 (1.3886) | | | | |
| | | | MSP | 65.8239 (1.5582) | | | | |
| | | | EBB | 66.6332 (1.4789) | | | | |
| | | | LCMV | 69.1971 (1.6150) | | | | |
| | | | eLORETA | 69.3565 (1.8551) | | | | |

**Supplementary Table 4: Comparative statistics for the combination of uncorrected MEG data and MRI z-scores**. Statistical summary of the N-way ANOVA assessing accuracy and AUC for the performance of combination MEG and MRI features across seven different classifiers, influenced by source localization techniques.

***Classifier model and inverse solution methods pairwise comparison: p value post hoc test (Tukey's HSD)
*** Pairwise comparison: p-value: Adjustment for multiple comparison: Bonferroni
*** Gaussian Naïve Bayes: GNB, Kernel Naïve Bayes (KNB), Wide Neural Network (WNN), Bilayered Neural Network (BNN), Ensemble: Subspace discriminant (ESD), Kernel Support Vector Machine (KSVM).

| Classification | Classifier | Source localization methods | Mean (SD) (%) | Contrast | p-value | 95% confidence interval | Pairwise comparison |
|---|---|---|---|---|---|---|---|
| Accuracy | Gaussian Naïve Bayes | Sensor | 67.9076 (1.0942) | GNB & KNB | < 0.001 | [-0.0518, -0.0466] | GNB < KNB |
| | | COH | 51.3333 (1.0452) | GNB & WNN | < 0.001 | [-0.1204, -0.1152] | GNB < WNN |
| | | MSP | 54.6799 (1.5614) | GNB & BNN | < 0.001 | [-0.0953, -0.0901] | GNB < BNN |
| | | EBB | 51.9538 (1.1672) | GNB & ESD | < 0.001 | [-0.0093, -0.0041] | GNB < ESD |
| | | LCMV | 67.5479 (0.8273) | GNB & KSVM | < 0.001 | [-0.1327, -0.1275] | GNB < KSVM |
| | | eLORETA | 64.6337 (1.0549) | GNB & GLMNET | < 0.001 | [-0.1467, -0.1415] | GNB < GLMNET |
| | Kernel Naïve Bayes | Sensor | 65.9241 (1.1144) | KNB & WNN | < 0.001 | [-0.0711, -0.0659] | KNB < WNN |
| | | COH | 63.0528 (1.4270) | KNB & BNN | < 0.001 | [-0.0461, -0.0409] | KNB < BNN |
| | | MSP | 63.7261 (1.3119) | KNB & ESD | < 0.001 | [0.0399, 0.0451] | KNB < ESD |
| | | EBB | 62.6007 (1.4040) | KNB & KSVM | < 0.001 | [-0.0834, -0.0782] | KNB < KSVM |
| | | LCMV | 68.6007 (0.8826) | KNB & GLMNET | < 0.001 | [-0.0975, -0.0923] | KNB < GLMNET |
| | | eLORETA | 63.6832 (1.1100) | WNN & BNN | < 0.001 | [0.0225, 0.0277] | WNN > BNN |
| | Wide Neural Network | Sensor | 69.2442 (1.8715) | WNN & ESD | < 0.001 | [0.1085, 0.1137] | WNN > ESD |
| | | COH | 71.9934 (1.7261) | WNN & KSVM | < 0.001 | [-0.0149, -0.0097] | WNN < KSVM |
| | | MSP | 70.4686 (1.5398) | WNN & GLMNET | < 0.001 | [-0.0289, -0.0237] | WNN < GLMNET |
| | | EBB | 71.8977 (1.7716) | BNN & ESD | < 0.001 | [0.0834, 0.0886] | BNN > ESD |
| | | LCMV | 73.8053 (1.6787) | BNN & KSVM | < 0.001 | [-0.0400, -0.0348] | BNN < KSVM |
| | | eLORETA | 71.3036 (1.8212) | BNN & GLMNET | < 0.001 | [-0.0540, -0.0488] | BNN < GLMNET |
| | Bilayered Neural Network | Sensor | 67.4917 (1.9844) | ESD & KSVM | < 0.001 | [-0.1259, -0.1208] | ESD < KSVM |
| | | COH | 69.7063 (1.9263) | ESD & GLMNET | < 0.001 | [-0.1400, -0.1348] | ESD < GLMNET |
| | | MSP | 68.2970 (2.1150) | KSVM & GLMNET | < 0.001 | [-0.0166, -0.0114] | KSVM < GLMNET |
| | | EBB | 69.2013 (2.0661) | Sensor & COH | < 0.001 | [0.0153, 0.0200] | Sensor > COH |
| | | LCMV | 70.9043 (1.8142) | Sensor & MSP | < 0.001 | [0.0158, 0.0204] | Sensor > MSP |
| | | eLORETA | 68.0726 (1.8238) | Sensor & EBB | < 0.001 | [0.0182, 0.0228] | Sensor > EBB |
| | Ensemble: Subspace discriminant | Sensor | 60.4026 (1.5106) | Sensor & LCMV | < 0.001 | [-0.0243, -0.0197] | Sensor < LCMV |
| | | COH | 59.6898 (1.0468) | COH & EBB | 0.006 | [0.0005, 0.0052] | COH > EBB |
| | | MSP | 60.8746 (1.1144) | COH & LCMV | < 0.001 | [-0.0420, -0.0373] | COH < LCMV |
| | | EBB | 59.9109 (1.1890) | COH & eLORETA | < 0.001 | [-0.0192, -0.0146] | COH < eLORETA |
| | | LCMV | 61.0462 (1.3138) | MSP & EBB | 0.036 | [9.6 x10⁻⁶, 0.0047] | MSP > EBB |
| | | eLORETA | 60.1551 (1.3930) | MSP & LCMV | < 0.001 | [-0.0424, -0.0378] | MSP < LCMV |
| | Kernel Support Vector Machine | Sensor | 70.7591 (1.5727) | MSP & eLORETA | < 0.001 | [-0.0197, -0.0150] | MSP < eLORETA |
| | | COH | 74.0495 (1.6098) | EBB & LCMV | < 0.001 | [-0.0448, -0.0402] | EBB < LCMV |
| | | MSP | 72.8350 (1.6315) | EBB & eLORETA | < 0.001 | [-0.0221, -0.0174] | EBB < eLORETA |
| | | EBB | 73.3564 (1.6031) | LCMV & eLORETA | < 0.001 | [0.0204, 0.0251] | LCMV > eLORETA |
| | | LCMV | 73.0429 (1.4396) | | | | |
| | | eLORETA | 72.0462 (1.6581) | | | | |
| | GLMNET | Sensor | 74.1419 (1.5582) | | | | |
| | | COH | 73.6898 (1.7875) | | | | |
| | | MSP | 72.3201 (1.4532) | | | | |
| | | EBB | 72.5875 (1.6437) | | | | |
| | | LCMV | 76.3135 (1.4706) | | | | |
| | | eLORETA | 75.4521 (1.4600) | | | | |
| AUC | Gaussian Naïve Bayes | Sensor | 70.8122 (0.9540) | GNB & KNB | < 0.001 | [-0.0450, -0.0398] | GNB < KNB |
| | | COH | 53.8450 (0.9639) | GNB & WNN | < 0.001 | [-0.1663, -0.1611] | GNB < WNN |
| | | MSP | 56.7324 (1.5209) | GNB & BNN | < 0.001 | [-0.1159, -0.1107] | GNB < BNN |
| | | EBB | 54.3628 (1.0891) | GNB & ESD | < 0.001 | [-0.0499, -0.0447] | GNB < ESD |
| | | LCMV | 72.1372 (0.7868) | GNB & KSVM | < 0.001 | [-0.1746, -0.1694] | GNB < KSVM |
| | | eLORETA | 67.6096 (1.0249) | GNB & GLMNET | < 0.001 | [-0.1795, -0.1743] | GNB < GLMNET |
| | Kernel Naïve Bayes | Sensor | 69.3242 (1.1712) | KNB & WNN | < 0.001 | [-0.1238, -0.1186] | KNB < WNN |
| | | COH | 64.2044 (1.3654) | KNB & BNN | < 0.001 | [-0.0735, -0.0683] | KNB < BNN |
| | | MSP | 64.5420 (1.1800) | KNB & ESD | < 0.001 | [-0.0074, -0.0022] | KNB < ESD |
| | | EBB | 64.1358 (1.4392) | KNB & KSVM | < 0.001 | [-0.1322, -0.1270] | KNB < KSVM |
| | | LCMV | 73.2903 (0.8476) | KNB & GLMNET | < 0.001 | [-0.1371, -0.1319] | KNB < GLMNET |
| | | eLORETA | 65.4641 (1.1331) | WNN & BNN | < 0.001 | [0.0477, 0.0529] | WNN > BNN |
| | Wide Neural Network | Sensor | 76.7638 (1.4800) | WNN & ESD | < 0.001 | [0.1138, 0.1190] | WNN > ESD |
| | | COH | 79.6185 (1.3694) | WNN & KSVM | < 0.001 | [-0.0109, -0.0057] | WNN < KSVM |
| | | MSP | 78.0150 (1.3025) | WNN & GLMNET (p = < 0.001, 95% C.I = | < 0.001 | [-0.0159, -0.0107] | WNN < GLMNET |
| | | EBB | 79.3387 (1.2513) | BNN & ESD | < 0.001 | [0.0635, 0.0687] | BNN > ESD |
| | | LCMV | 81.8735 (1.2447) | BNN & KSVM | < 0.001 | [-0.0613, -0.0561] | BNN < KSVM |
| | | eLORETA | 78.0918 (1.3997) | BNN & GLMNET | < 0.001 | [-0.0662, -0.0610] | BNN < GLMNET |
| | Bilayered Neural Network | Sensor | 72.0099 (2.0213) | ESD & KSVM | < 0.001 | [-0.1273, -0.1221] | ESD < KSVM |
| | | COH | 74.6975 (2.1610) | ESD & GLMNET | < 0.001 | [-0.1323, -0.1271] | ESD < GLMNET |
| | | MSP | 73.1533 (2.3070) | KSVM & GLMNET | < 0.001 | [-0.0075, -0.0023] | KSVM < GLMNET |
| | | EBB | 74.0419 (2.0196) | Sensor & COH | < 0.001 | [0.0226, 0.0273] | Sensor > COH |
| | | LCMV | 76.7516 (1.7386) | Sensor & MSP | < 0.001 | [0.0279, 0.0325] | Sensor > MSP |
| | | eLORETA | 72.8458 (1.8177) | Sensor & EBB | < 0.001 | [0.0267, 0.0314] | Sensor > EBB |
| | Ensemble: Subspace discriminant | Sensor | 69.8475 (1.9510) | Sensor & LCMV (p = < 0.001, 95% C.I = | < 0.001 | [-0.0319, -0.0273] | Sensor < LCMV |
| | | COH | 65.6420 (1.4264) | Sensor & eLORETA | < 0.001 | [0.0050, 0.0096] | Sensor > eLORETA |
| | | MSP | 65.3532 (1.1564) | COH & MSP | < 0.001 | [0.0029, 0.0075] | COH > MSP |
| | | EBB | 65.3560 (1.5920) | COH & EBB | < 0.001 | [0.0018, 0.0064] | COH > EBB |
| | | LCMV | 70.4326 (1.6402) | COH & LCMV | < 0.001 | [-0.0569, -0.0522] | COH < LCMV |
| | | eLORETA | 67.2213 (1.5177) | COH & eLORETA | < 0.001 | [-0.0200, -0.0153] | COH < eLORETA |
| | Kernel Support Vector Machine | Sensor | 78.1569 (1.4238) | MSP & LCMV | < 0.001 | [-0.0621, -0.0574] | MSP < LCMV |
| | | COH | 80.7324 (1.4665) | MSP & eLORETA | < 0.001 | [-0.0252, 00.0206] | MSP < eLORETA |
| | | MSP | 79.6859 (1.4019) | EBB & LCMV | < 0.001 | [-0.0610, -0.0563] | EBB < LCMV |
| | | EBB | 80.3710 (1.4818) | EBB & eLORETA | < 0.001 | [-0.0241, -0.0194] | EBB < eLORETA |
| | | LCMV | 80.6503 (1.4610) | LCMV & eLORETA | < 0.001 | [0.0345, 0.0392] | LCMV > eLORETA |
| | | eLORETA | 79.1000 (1.5020) | | | | |
| | GLMNET | Sensor | 79.9510 (1.6875) | | | | |
| | | COH | 80.6548 (1.8568) | | | | |
| | | MSP | 78.2596 (1.7074) | | | | |
| | | EBB | 78.9229 (1.9890) | | | | |
| | | LCMV | 82.4386 (1.4350) | | | | |
| | | eLORETA | 81.4321 (2.0431) | | | | |